\documentclass{article}

\PassOptionsToPackage{table}{xcolor}

\usepackage{iclr2027_conference,times}
\iclrfinalcopy %

\usepackage[utf8]{inputenc} %
\usepackage[T1]{fontenc}    %
\usepackage{hyperref}       %
\usepackage{url}            %
\usepackage{booktabs}       %
\usepackage{amsfonts}       %
\usepackage{nicefrac}       %
\usepackage{microtype}      %
\usepackage{xcolor}         %

\usepackage{algorithm}
\usepackage{algorithmic}

\usepackage{amsmath}
\usepackage{amssymb}
\usepackage{mathtools}
\usepackage{amsthm}

\usepackage{bm,bbm}
\usepackage{xspace}
\usepackage{multirow}
\usepackage{cancel}
\usepackage[normalem]{ulem}
\usepackage{enumitem}
\usepackage{caption}
\usepackage{booktabs}

\usepackage{subcaption}
\usepackage{graphicx}
\usepackage{threeparttable, array, float}
\usepackage{wrapfig}

\usepackage{tabularx}
\usepackage[table]{xcolor}
\usepackage{array}
\usepackage{booktabs}
\usepackage{threeparttable}
\usepackage{subcaption}
\usepackage{pifont}
\usepackage{soul}
\usepackage{circledsteps}

\usepackage[capitalize,noabbrev]{cleveref}

\usepackage{tcolorbox}

\newtheorem{theorem}{\bf Theorem}
\newtheorem{lemma}{\bf Lemma}
\newtheorem{corollary}{\bf Corollary}

\newtheorem{fact}{\bf Fact}
\theoremstyle{remark}

\newtheorem{assumption}{\bf Assumption}

\crefname{assumption}{Assumption}{Assumptions}
\Crefname{assumption}{Assumption}{Assumptions}

\Crefname{condition}{Condition}{Conditions}
\Crefname{proposition}{Proposition}{Proposition}

\usepackage{tikz}
\usepackage{amsmath,amssymb}
\usetikzlibrary{arrows.meta,positioning,calc}

\title{Finite-Time Convergence and Regret Guarantees for Neural Reinforcement Learning in POMDPs}

\author{Jingjuan Huang$^{1}$, Zhiyao Zhang$^{1}$, Alvaro Velasquez$^{2}$, Nathaniel D. Bastian$^{3}$, Jia Liu$^{1}$ \\
$^{1}$Ohio State University \qquad $^{2}$University of Colorado at Boulder \qquad $^{3}$Johns Hopkins University}

\begin{document}

\maketitle
\lhead{Preprint} %

\begin{abstract}
Reinforcement learning (RL) in Partially Observable Markov Decision Processes (POMDPs) is computationally challenging due to the curse of long-range dependence on history. 
Fortunately, recent studies have shown that a range of POMDPs that satisfy the filter stability assumption can be well approximated by a superstate Markov Decision Process (MDP) based on a truncated history in a finite-memory window, with a value gap that decays with the window length.
However, existing finite-time convergence analyses on such superstate MDP surrogates are either limited to over-simplified linear function approximation or rely on an i.i.d. sampling oracle that is violated by POMDP trajectories, both of which severely limit the applicability of these theoretical results in practice.
This motivates us to bridge the gap by investigating RL in POMDPs with {\em nonlinear} neural value function approximation. 
Our main contributions are threefold:
i) We first establish a new finite-state MDP {\em point-wise} approximation error bound, which lays the foundation for our RL algorithm design and analysis in POMDPs;
ii) By applying Temporal-Difference (TD) learning to POMDP-generated superstate transitions and targeting the Bellman fixed point of the superstate MDP, we prove a finite-time critic error bound with nonlinear neural value function approximation;
and iii) By combining the neural critic function approximation error bound with the policy mirror descent framework, we obtain a finite-time sublinear regret against the optimal POMDP value.
To the best of our knowledge, such a finite-time sublinear regret based on neural function approximation is the first in the literature for finite-memory POMDPs.
Our numerical experiments also verify our theoretical results. 
\end{abstract}

\section{Introduction}
\label{sec:intro}
Reinforcement learning (RL) has achieved enormous empirical success in fully observable environments, yet many real-world decision problems suffer from partial observability of the underlying state, such as robots operating under noisy sensing \citep{kurniawati2021partially}, autonomous vehicles having to reason about hidden intentions of surrounding agents \citep{sunberg2022improving}, and clinical decision-making relying on indirect measurements of patient states \citep{hauskrecht2000planning}. 
More recently, agentic AI systems have also been modeled as partially observable problems \citep{zhang2025landscape}. 
Thus, it is more appropriate to model such problems as partially observable Markov decision processes (POMDPs) \citep{kaelbling1998planning}, and the corresponding reinforcement learning problem in POMDPs is known as partially observable reinforcement learning (PORL).

In POMDPs, the current observation is generally insufficient to infer the underlying current state.
Therefore, the agent needs to rely on the full action-observation history. 
A standard approach for solving POMDPs is to maintain a \emph{belief state} \citep{kaelbling1998planning}, which is a sufficient statistic of the history. 
However, the belief state lies in a continuous simplex, 
and exact belief updates require known transition and observation models,
making exact belief-based methods computationally intractable. 
Practical algorithms therefore approximate the belief state through either a recurrent representation of the history \citep{cayci2024recurrent} or a finite memory \citep{kara2023convergence,anjarlekar2025scalable}. 
Although recurrent representations are more expressive, training them over long time horizons could suffer from vanishing or exploding gradients \citep{pascanu2013difficulty}. 
In contrast, finite-memory representations act directly on a fixed-length window, leading to simpler and more parallelizable training. 
Empirically, finite-memory baselines such as simple frame stacking have been shown to be competitive memory models on partially observable benchmarks~\citep{morad2023popgym}. Also, \citet{hausknecht2015deep} found no systematic advantage of recurrence over finite-memory on Atari-style benchmarks.
Beyond these practical motivations, truncating the history into a finite-memory window yields a so-called \emph{superstate}, which facilitates an explicit accuracy--memory trade-off through the window length and induces a tractable finite-state surrogate MDP.

Given the finite-state surrogate MDP, one can potentially solve RL problems in POMDPs by using RL algorithmic frameworks developed for conventional MDP (e.g., the Policy Mirror Descent (PMD) framework
\citep{abbasi2019politex,xiao2022convergence, lan2023policy, zhan2023policy}
, which further unifies a wide range of algorithms including natural policy gradient (NPG), trust-region policy optimization (TRPO), and proximal policy optimization (PPO) \citep{tomar2020mirror, alfano2023novel}). 
However, existing finite-time analyses for these RL algorithms in PORL have two limitations:
\vspace{-.2in}
\begin{list}{\labelitemi}{\leftmargin=1.5em \itemindent=-0.0em \itemsep=.1em}
\item[1)] In finite-memory methods, most existing works adopt linear function approximation.
For example, \citet{cayci2024finite} analyzed NAC algorithm with a finite-state controller, and \citet{anjarlekar2025scalable} analyzed PMD algorithm on the superstate MDP. However, it is well-known that linear models' expressiveness is limited and do not offer universal approximation no matter how many linear layers are stacked.
When a POMDP value function is not well approximated by linear models, the bias manifests as a persistent error term in the theoretical analysis. 

\item[2)] In contrast, neural function approximation enjoys the powerful universal approximation property (i.e.,  neural networks can approximate broad classes of functions arbitrarily well by scaling up the network width and depth \citep{hornik1989multilayer,shen2022optimal}). 
However, existing works on neural function approximation for POMDPs assume i.i.d. sampling from the discounted visitation distribution (e.g., \citet{cayci2024recurrent}). 
Such an i.i.d. sampling assumption does not hold in on-policy POMDP trajectories, where samples are long-range temporal dependent.
\end{list}

In light of the growing importance of PORL in many emerging applications and the above limitations in its theoretical understanding, the following fundamental open question naturally arises: 

\begin{tcolorbox}[left=1.2pt,right=1.2pt,top=1.2pt,bottom=1.2pt]
\textbf{(Q)}: Could we establish a rigorous theoretical understanding of solving POMDPs via reinforcement learning with {\em neural} value function approximation through a superstate surrogate MDP?
\end{tcolorbox}

Answering the above question is highly non-trivial due to the following fundamental challenges:
\vspace{-.05in}
\begin{itemize}[leftmargin=*, topsep=2pt, itemsep=2pt, parsep=0pt]
\item \textbf{Model Mismatch:} In PORL through superstate MDPs, the neural critic is updated with data sampled from POMDP trajectories, whereas the Bellman fixed point is defined on the superstate MDP. 
This discrepancy is absent in conventional temporal RL critic analyses, where the sampling distribution and the Bellman target are defined on the same MDP \citep{cai2019neural}. Establishing a finite-time convergence guarantee for neural critics in PORL therefore requires a new analysis.
    
\item \textbf{Non-i.i.d. Sampling within Iterations and Non-stationarity across Iterations.} Within each iteration, the neural critic is updated along a POMDP trajectory rather than i.i.d. samples. 
Standard Markovian temporal difference (TD) analyses \citep{bhandari2018finite, cai2019neural} compare an MDP trajectory with the stationary distribution of the same underlying Markov chain. 
In PORL, however, the POMDP trajectory does {\em not} follow the superstate MDP kernel. 
Exacerbating the problem is the fact that, across iterations, the discounted visitation distribution is {\em non-stationary} and changes with the policy.
Thus, the critic error established under the current visitation needs to be updated to that of the optimal policy, which introduces another layer of difficulty.
\end{itemize}

Our key contribution in this paper is that we address these two fundamental challenges by demonstrating that standard TD learning for critics with {\em neural} function approximation, when coupled with the PMD framework for superstate MDPs, achieves {\em sublinear regret} against the optimal POMDP value.
Our main results are summarized as follows and organized in the roadmap as shown in Fig.~\ref{fig:roadmap}:

\begin{itemize}[leftmargin=*, topsep=2pt, itemsep=2pt, parsep=0pt]
\item \textbf{Point-wise POMDP-to-Superstate MDP Approximation:}
We first establish a {\em pointwise} value approximation bound between the POMDP and the induced superstate MDP under a filter-stability assumption (Step~\Circled{1} in Fig.~\ref{fig:roadmap}).
We also show that the reward-function and transition-kernel mismatches between the POMDP and superstate MDP models can be controlled by the same filter-stability error. This identifies the key mechanism behind the superstate MDP approximation and provides the foundation for our learning algorithm.
  
\item \textbf{Finite-Time Neural TD Evaluation under POMDP Trajectory Sampling:}
We establish a finite-time error bound showing that the neural critic trained from POMDP trajectories could well approximate the Bellman fixed point of the induced superstate MDP (Step~\Circled{2} in Fig.~\ref{fig:roadmap}).
To handle the model mismatch and non-i.i.d. trajectory sampling, we develop a {\em new} Lyapunov descent analysis that separates trajectory-sampling error, neural-model error, and partial-observability error.

\item \textbf{Sublinear Regret for Neural PMD Algorithm:}
We integrate the critic guarantee into the PMD analysis (Step~\Circled{3} in Fig.~\ref{fig:roadmap}) and prove sublinear regret for the overall algorithm (Step~\Circled{4} in Fig.~\ref{fig:roadmap}). 
To address the non-stationarity across policy iterations, we bridge this distribution mismatch through a concentrability assumption, and show that the regret bound explicitly separates the effects of neural critic error, finite-memory approximation error and policy optimization error.

\end{itemize}
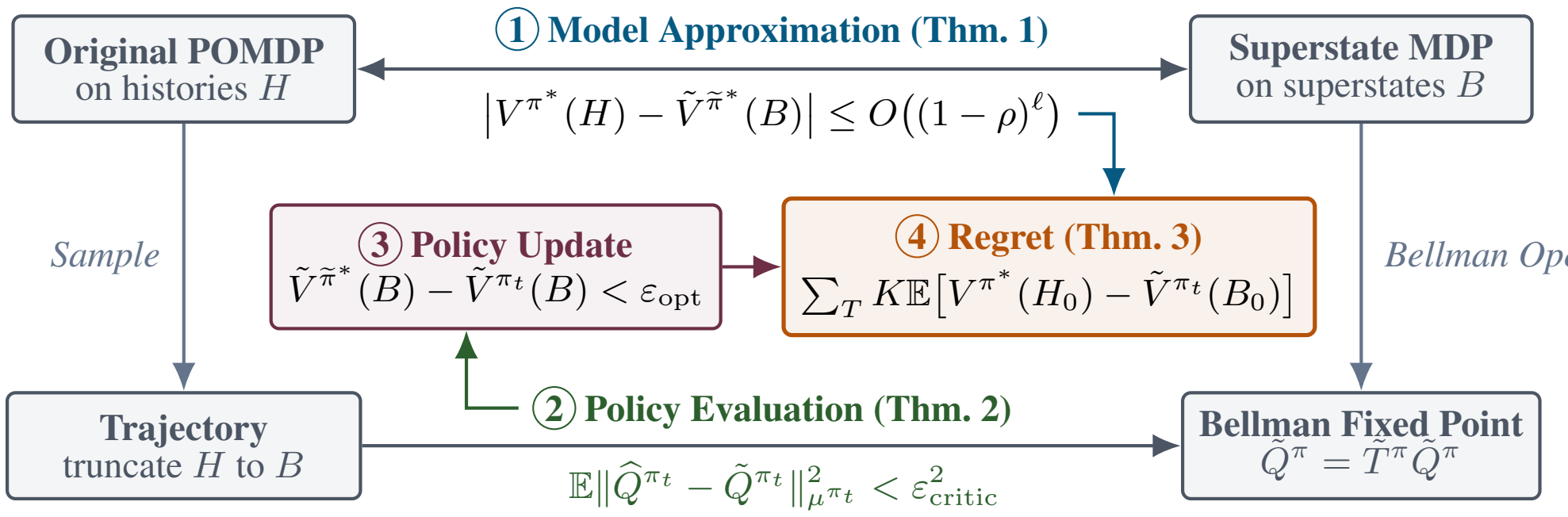
\begin{figure}[t!]
\centering

\definecolor{cPOMDP}{HTML}{1E2761}
\definecolor{cSMDP}{HTML}{065A82}
\definecolor{cTRAJ}{HTML}{B85042}
\definecolor{cBELL}{HTML}{2C5F2D}
\definecolor{cPMD}{HTML}{6D2E46}
\definecolor{cTD}{HTML}{0D9488}
\definecolor{cRN}{HTML}{B45309}
\definecolor{cMUTED}{HTML}{64748B}
\definecolor{cBOX}{HTML}{4B5563}

\resizebox{\linewidth}{!}{%
\begin{tikzpicture}[
  >={Latex[length=2.4mm,width=2mm]},
  font=\small,
  corner/.style={draw,line width=0.8pt,rounded corners=2pt,align=center,inner sep=4pt},
  neutral/.style={corner,draw=cBOX,fill=cBOX!6,text=cBOX,minimum width=2.9cm,minimum height=0.9cm},
  neutralwide/.style={corner,draw=cBOX,fill=cBOX!6,text=cBOX,minimum width=3.0cm,minimum height=0.9cm},
  pmd/.style={corner,draw=cPMD,fill=cPMD!6,text=cPMD,minimum width=3.65cm,minimum height=1.05cm},
  regret/.style={corner,draw=cRN,fill=cRN!8,text=cRN,line width=1pt,minimum width=3.65cm,minimum height=1.05cm},
  edgelbl/.style={font=\footnotesize\itshape,text=cMUTED},
]

\node[neutral] (P) at (0,0)
  {\textbf{Original POMDP}\\[-1pt]\footnotesize on histories $H$};

\node[neutral] (S) at (10,0)
  {\textbf{Superstate MDP}\\[-1pt]\footnotesize on superstates $B$};

\node[neutralwide] (T) at (0,-3.2)
  {\textbf{Trajectory}\\[-1pt]\footnotesize truncate $H$ to $B$};

\node[neutralwide] (B) at (10,-3.2)
  {\textbf{Bellman Fixed Point}\\[-1pt]\footnotesize $\tilde Q^\pi=\tilde T^\pi\tilde Q^\pi$};

\node[pmd] (PMD) at (2.65,-1.68) {%
{\small\textbf{\Circled{3} Policy Update}}\\[-1pt]
{\footnotesize{\color{black}$\tilde V^{\widetilde\pi^*}(B)-\tilde V^{\pi_t}(B)<\varepsilon_{\mathrm{opt}}$}}
};

\node[regret] (R) at (7.35,-1.68) {%
{\small\textbf{\Circled{4} Regret (Thm.~\ref{thm:regret})}}\\[2pt]
{\footnotesize{\color{black}$\sum_T K\mathbb E\bigl[V^{\pi^*}(H_0)-\tilde V^{\pi_t}(B_0)\bigr]$}}
};

\draw[<->,thick,cBOX] (P.east) -- (S.west)
  coordinate[midway] (FS);

\node[font=\bfseries\small,text=cSMDP] (MA) at ($(FS)+(0,0.34)$)
  {\Circled{1} Model Approximation (Thm.~\ref{thm:finite-memory-near-optimality})};

\node[font=\footnotesize,text=cSMDP] (FSform) at ($(FS)+(0,-0.38)$)
  {{\color{black}$\bigl|V^{\pi^*}(H)-\tilde V^{\widetilde\pi^*}(B)\bigr|\le O\bigl((1-\rho)^\ell\bigr)$}};

\draw[->,thick,cMUTED] (P.south) -- (T.north)
  node[midway,left=2pt,edgelbl]{Sample};

\draw[->,thick,cMUTED] (S.south) -- (B.north)
  node[midway,right=2pt,edgelbl]{Bellman Operator};

\draw[->,thick,cBOX] (T.east) -- (B.west);

\node[font=\bfseries\small,text=cBELL] (TDlabel) at (5.0,-2.88)
  {\Circled{2} Policy Evaluation (Thm.~\ref{thm:critic-error})};

\node[font=\footnotesize,text=cBELL] (TDform) at (5.1,-3.54)
  {$\mathbb E\|\widehat Q^{\pi_t}-\tilde Q^{\pi_t}\|_{\mu^{\pi_t}}^2<\varepsilon_{\mathrm{critic}}^2$};

\coordinate (TDx) at (2.4,0);
\draw[->,thick,cBELL] (TDlabel.west) -- (TDx |- TDlabel.west) -- (TDx |- PMD.south);

\draw[->,thick,cPMD] (PMD.east) -- (R.west);

\coordinate (FSx) at (7.9,0);
\draw[->,thick,cSMDP] (FSform.east) -- (FSx |- FSform.east) -- (FSx |- R.north);

\end{tikzpicture}
}%
\vspace{-.05in}
\caption{Roadmap for our theoretical finite-time analysis of reinforcement learning in POMDPs with neural value function approximation.}
\vspace{-.2in}
\label{fig:roadmap}
\end{figure}

\section{Related Work}

In this section, we provide an overview of three lines of research that are closely related to this work, thereby placing our contributions into a comparative perspective.

\textbf{1) Finite-memory Learning for POMDPs:}
Since exact belief-state methods are intractable for POMDPs \citep{smallwood1973optimal,lovejoy1991survey, kaelbling1998planning}, various practical surrogates have been proposed based on recurrent representations \citep{hausknecht2015deep, cayci2024recurrent} or finite-memory controllers and sliding-window superstates
\citep{poupart2003bounded, amato2012optimizing, meuleau2013learning, kara2022near, kara2023convergence,anjarlekar2025scalable}.
A central tool for the latter is \emph{filter stability}, which controls the gap between distinct histories that map to the same superstate. 
Exponential filter stability can be ensured under $\gamma$-observability \citep{golowich2023planning, cai2024provable}, or via Dobrushin coefficients of the transition and observation kernels \citep{mcdonald2020exponential}. The closest work to ours is \citet{anjarlekar2025scalable}, which analyzes policy-based learning on a superstate MDP with linear function approximation. 
We adopt the same finite-memory perspective, and further investigate neural network approximation under milder assumptions compared to \citet{anjarlekar2025scalable}.

\textbf{2) Neural Function Approximation for RL in MDPs:}
For fully observable MDPs, finite-time guarantees have been developed for neural value-based and policy-gradient methods \citep{cai2019neural,sun2022finite,gaur2024closing, zhang2025finite}.
In POMDPs, neural methods typically aggregate histories through recurrent or latent-state representations
\citep{wierstra2007solving,hausknecht2015deep,igl2018deep}.
Recent finite-time analyses have studied recurrent natural actor-critic methods under an i.i.d. sampling assumption
\citep{cayci2024recurrent}. However, finite-time analysis for neural approximation in POMDPs under standard on-policy 
trajectory sampling 
remains open.

\textbf{3) Policy Mirror Descent (PMD) and Policy Gradient Methods:}
A growing line of work casts NPG, TRPO, and PPO as instances of policy mirror descent \citep{kakade2001natural,schulman2015trust,schulman2017proximal,shani2020adaptive, tomar2020mirror, alfano2023novel}. 
For fully observable MDPs, global convergence and finite-time guarantees have been established for tabular softmax methods \citep{mei2020global,cen2022fast}, regularized policy mirror descent
\citep{lan2023policy,zhan2023policy}, and broader parametric policy classes, including log-linear or general
parameterizations \citep{agarwal2021theory,yuan2022linear,alfano2023novel}. However, the finite-time behavior of PMD with neural critic approximation remains unexplored in POMDP settings.

\section{Problem Statement and Preliminaries}
\label{Preliminaries}
\textbf{1) POMDP Model:}
We consider a POMDP $(\mathcal S,\mathcal O,\mathcal A,P,\Phi,r,\gamma)$ with finite state space $\mathcal S$, finite observation space $\mathcal O$, finite action space $\mathcal A$, transition probabilities $P:\mathcal{S}\times\mathcal{A}\rightarrow P(\mathcal{S})$
, observation probabilities $\Phi(o\mid s)$, reward function $r:\mathcal S\times\mathcal A\to[0,r_{\max}]$, and discount factor $\gamma\in(0,1)$. The initial state $s_0\in\mathcal{S}$ is sampled from a distribution $\mathcal D$.
The dynamics satisfy:
\begin{align*}
    & \mathbb{P}(S_{t+1} = s' |S_t = s, A_t = a) = P(s'|s,a), & \forall t=0,1, \dotsc, \\
    & \mathbb{P}(O_{t+1} = o | S_{t+1} = s') = \Phi(o|s'), & \forall t=0,1, \dotsc.
\end{align*}
Since the latent state process is unobserved, the agent acts on the history $H_{t+1} = (H_{t}, A_{t},O_{t+1})\in\mathbb{H}$ with $H_0 = \varnothing$. 
The \emph{belief state} $b_H(s):=\mathbb P(S_t=s\mid H_t=H)$ is a sufficient statistic of $H$.

\textbf{2) Superstate MDP:}
To construct a superstate MDP, we fix a window length $\ell\in\mathbb N$ and let
$\mathcal G:\mathbb H\to\mathsf B$ map each history to its most recent
$\ell$ action--observation pairs:
$B_t:=\mathcal G(H_t)=\bigl(A_{\max\{0,t-\ell\}},O_{\max\{1,t-\ell+1\}},\dots,A_{t-1},O_t\bigr),$
with $\mathsf B:=\{\mathcal G(H):H\in\mathbb H\}$ being the resulting finite
superstate space.  
Following \citet{anjarlekar2025scalable}, given a belief $b\in\Delta(\mathcal S)$, an action $a$, and an observation $o$, we define the belief-state update operator $\mathcal U_{a,o}$ using Bayes' rule as follows:
\[
\left[\mathcal U_{a,o}(b)\right](s')
=
\frac{
\Phi(o\mid s')\sum_{s\in\mathcal S}P(s'\mid s,a)b(s)
}{
\sum_{\bar s\in\mathcal S}\Phi(o\mid\bar s)
\sum_{s\in\mathcal S}P(\bar s\mid s,a)b(s)
}.
\]
Set $B_0=\varnothing$ and $b_{B_0}=\mathcal D$. For $t\geq 1$, define $b_{B_t}=\mathcal U_{A_{t-1},O_t}\circ\cdots\circ\mathcal U_{A_{\max\{0,t-\ell\}},O_{\max\{1,t-\ell+1\}}}(\mathcal D)$. We write $b_B$ for the belief associated with superstate $B$.
The superstate MDP
$(\mathsf B,\mathcal A,\widetilde{\mathcal P},\tilde r,\gamma)$ has
reward and transition kernel defined as follows:
\[
\resizebox{\linewidth}{!}{$\displaystyle
\begin{aligned}
&\tilde{r}(B,a)
\!=\!\! {\textstyle\sum_s} b_B(s)\,r(s,a), \,\,
\widetilde{\mathcal{P}}(B' | B,a)
\!=\!\!{\textstyle\sum_{o,s,s'}} \mathbf{1}[\mathcal{G}(B\!\parallel\!\{a,o\})=B']\,\Phi(o\mid s')\,P(s'\mid s,a)\,b_B(s),
\end{aligned}
$}
\]
where $\mathbf{1}[\cdot]$ is the indicator function, and $B\parallel\{a,o\}$ denotes the concatenation of the action-observation pair $\{a,o\}$ to the current superstate $B$.
For any $B \in \mathsf B$ and any policy $\pi(\cdot | B)$ on the superstate MDP, the value function and action-value function are defined as 
\[
\resizebox{\linewidth}{!}{$\displaystyle
\begin{aligned}
&\tilde V^\pi(B)=\mathbb E\big[{\textstyle\sum_{t=0}^{\infty}}\gamma^t \tilde r(B_t,a_t)\,\bigm|\, B_0=B\big], \ 
\tilde Q^\pi(B,a)=\mathbb E\big[{\textstyle\sum_{t=0}^{\infty}}\gamma^t \tilde r(B_t,a_t)\,\bigm|\, B_0=B,\ a_0=a\big],
\end{aligned}
$}
\]
and the Bellman operator can be written as:
\[
  (\tilde T^\pi\tilde Q)(B,a)
  :=\tilde r(B,a)
  +\gamma{\textstyle\sum_{B'\in\mathsf B}\sum_{a'\in\mathcal A}}
    \widetilde{\mathcal P}(B'\mid B,a)\,\pi(a'\mid B')\,
    \tilde Q(B',a').
\]
And the true Q-function $\tilde Q^\pi$ is the unique Bellman fixed point satisfying $\tilde Q^\pi=\tilde T^\pi(\tilde Q^\pi)$.

\textbf{3) Neural Function Approximation:}
We parameterize the critic $\widehat Q(B,a;W)$ by a
$d$-layer ReLU network of width~$m$.  For each superstate-action pair $(B,a)\in\mathsf B\times\mathcal A$, let
$x=\psi(B,a)\in\mathbb R^p$ with $\|x\|_2=1$ be a one-to-one feature
encoding.  The neural network is parameterized as:
\[
x^{(0)}=Ux,\quad x^{(h)}=\frac{1}{\sqrt m}\,\mathrm{ReLU}\!\left(W^{(h)}x^{(h-1)}\right)\ \text{for each}\ h\in[d],\quad \widehat Q(x;W)=c^\top x^{(d)},
\]
where $[d]:=\{1,\dots,d\}$, $U\in\mathbb R^{m\times p}$, $W^{(h)}\in\mathbb R^{m\times m}$, and $c\in\mathbb R^m$ are the network parameters, and $\mathrm{ReLU}(u):=\max\{u,0\}$ is the activation function.
For convenience, we write $W=\big(\mathrm{vec}(W^{(1)})^\top,\dots,\mathrm{vec}(W^{(d)})^\top\big)^\top\in\mathbb R^{dm^2}$.
At initialization, entries of $U$ and $\{W^{(h)}\}_{h=1}^d$ are independently sampled from $\mathcal N(0,2)$ and entries of $c$ are independently sampled from $\mathcal N(0,1)$.
During training, we only update $W$, while keeping $U$ and $c$ fixed at their random initialization. 
It is worth noting that the bias-free notation is adopted for simplicity. Without loss of generality, affine ReLU layers with bias can be handled by augmenting the input and each hidden representation with a fixed coordinate, thereby absorbing the bias terms into the weight matrices.
We also note that the use of ReLU activation here is only for illustrative purposes and our theoretical results extend to many common activations (e.g., Sigmoid and Tanh) with the same approximation order up to constant width and depth factors~\citep{zhang2024deep}.

\section{Theory and Algorithm Design for Solving POMDPs via Superstate MDP}

In this section, we present our theoretical results and algorithmic designs for solving POMDPs via a superstate MDP with neural value function approximation.

\subsection{A New Point-wise POMDP-to-Superstate MDP Approximation Bound}
We first quantify the error introduced by replacing the POMDP with the finite-memory superstate MDP by showing a {\em new} point-wise approximation bound between them. 
Toward this end, we first state two assumptions that are needed to establish the point-wise approximation bound.
\begin{assumption}[Bounded Reward]
\label{assump:bounded-reward}
There exists a constant $r_{\max}>0$ such that for all $(s,a)\in\mathcal S\times\mathcal A$,
$
|r(s,a)|\le r_{\max}.
$
\end{assumption}

\begin{assumption}[Filter Stability]
\label{assump:filter-stability}
There exists a constant \(\rho\in(0,1)\) such that for any two histories \(H\) and \(H'\) satisfying \(\mathcal G(H)=\mathcal G(H')\), the corresponding belief states satisfy
$
\|b_H-b_{H'}\|_{\mathrm{TV}}\le (1-\rho)^\ell.
$
\end{assumption}
Assumption~\ref{assump:filter-stability} is a point-wise form of exponential filter stability. 
Intuitively, the influence of observations and actions prior to the last $\ell$ steps decays at a geometric rate. 
This condition controls the mismatch between the true POMDP history process and superstate MDP.
Noticing that the \textit{Uniform Filter Stability Condition} in \citet{anjarlekar2025scalable} is a sufficient condition for 
Assumption~\ref{assump:filter-stability}. 
Hence, our assumption is {\em weaker.}
Also, in the literature, the filter-stability assumption has been verified in numerous practical POMDP applications (e.g., software-reliability tracking \citep{debrabant2013stability}, robot localization \citep{lourenco2018globally}, and neural spike-train models \citep{whiteley2021dimension}).

We now state the resulting POMDP-to-Superstate MDP approximation guarantee,
which bounds the value gap between the POMDP and the induced superstate MDP, thereby justifying the superstate MDP as an appropriate finite-memory surrogate used by our learning algorithm.

\begin{theorem}[Finite-Memory Near-Optimality]
\label{thm:finite-memory-near-optimality}
Under Assumptions~\ref{assump:bounded-reward} and~\ref{assump:filter-stability}, for any history $H\in\mathbb H$, let $B=\mathcal G(H)$.
Let $\pi^*$ be an optimal policy of the original POMDP among all unrestricted history-dependent policies.
Let $\widetilde\pi^*$ be an optimal policy of the induced superstate MDP.
It then holds that
$\|V^{\pi^*}(H)-\tilde V^{\widetilde\pi^*}(B)\|\le\frac{2r_{\max}}{(1-\gamma)^2}(1-\rho)^\ell$.
\end{theorem}

Theorem~\ref{thm:finite-memory-near-optimality} says that the approximation bound is point-wise controlled by the window length $\ell$ for any history $H$.
Intuitively, under filter stability, histories with the same superstate induce close beliefs. Hence, the corresponding reward-function and transition-kernel mismatches also decay at the rate of $O((1-\rho)^\ell)$.
This is the fundamental reason why the superstate MDP approximates the POMDP.
See Appendix~\ref{app:optimality-bridge} for proof details.
Our result is in the same spirit as those in \citep{kara2023convergence} and \citep{anjarlekar2025scalable} and yet with key differences.
Specifically, compared to our point-wise result, the approximation bound in \citep{kara2023convergence} is in a weaker expectation sense.
Meanwhile, with a stronger assumption, a similar point-wise bound with a better dependence on the $(1-\gamma)$-factor is proved in \citep{anjarlekar2025scalable}. 
Note that our bound is already sufficient for proving a sublinear regret, which is infeasible with linear function approximation adopted in \citep{anjarlekar2025scalable}.

\subsection{Policy Optimization Algorithm Design through the Lens of Superstate MDP}
\label{sec:policy-opt-design}
Based on the superstate MDP approximation, our goal is to develop an RL algorithm with neural function approximation to learn a near-optimal policy in the sense of achieving sublinear regret. 
We consider a two-level neural policy iteration framework consisting of: (i) a neural TD critic (Algorithm~\ref{alg:neural_superstate_td})  for estimating Bellman fixed point $\tilde Q^\pi$ of superstate MDP; and (ii) a policy update step (Algorithm~\ref{alg:superstate_policy_opt}) based on a closed-form exponential-weights update on the action simplex.
Within each policy evaluation phase, training samples are collected from a single on-policy POMDP trajectory.

\textbf{1) Policy Evaluation:}
We consider restart sampling: $(S,H,B)\leftarrow(S_0\sim\mathcal D,\varnothing,\varnothing)$ with probability $1-\gamma$, otherwise the sampler follows the original kernel $P$.
Then we adopt the standard semi-gradient TD-learning for policy evaluation in Algorithm~\ref{alg:neural_superstate_td}.
For a given policy $\pi$, our goal is to find the best neural network parameters \(W\in\mathcal{B}\) (defined in Algorithm~\ref{alg:neural_superstate_td}) such that \(\widehat Q(\cdot,\cdot;W)\) approximates the Bellman fixed point \(\tilde Q^\pi\) of the superstate MDP.
At step $k$, if no reset occurs, then
the agent observes the current superstate $B_k = \mathcal G(H_k)$, selects $a_k \sim \pi(\cdot \mid B_k)$, and the underlying POMDP returns reward \(r_k\) and the next observation \(o_{k+1}\).
The next superstate is obtained by concatenation and truncation: 
$H_{k+1}=H_k\parallel(a_k,o_{k+1})$ and $B_{k+1}=\mathcal G(H_{k+1})$,
followed by $a_{k+1}\sim\pi(\cdot\mid B_{k+1})$. The critic is updated by TD learning on the transition $(B_k, a_k, r_k, B_{k+1}, a_{k+1})$, with the average parameter $\bar{W}_K$ taken as the output.
If a reset occurs at step $k$, the critic parameters remain unchanged and the next action is sampled from $\pi(\cdot\mid B_0)$.

\textbf{2) Policy Update:}
Algorithm~\ref{alg:superstate_policy_opt} starts from the uniform policy and performs $T$ outer policy updates. 
In each outer iteration $t$, we run Algorithm~\ref{alg:neural_superstate_td} under the current policy $\pi_t$ for $K$ critic steps, obtaining the averaged critic $\widehat Q_t(B,a)$.
The policy is then updated independently in each superstate by the KL-regularized mirror-descent step as follows:
\[
\pi_{t+1}(\cdot\mid B)
:= \operatorname*{argmin}_{p\in\Delta(\mathcal A)}
\Bigl\{
-\kappa\,\langle p,\widehat Q_t(B,\cdot)\rangle
+ D_{\mathrm{KL}}\bigl(p \,\|\, \pi_t(\cdot\mid B)\bigr)
\Bigr\},
\]
which admits an exponential-weights closed-form update in Algorithm~\ref{alg:superstate_policy_opt} (see Appendix~\ref{policy improvement} for details). 

\newsavebox{\algorithmTrialCriticBox}
\newsavebox{\algorithmTrialPolicyBox}
\newlength{\algorithmTrialLeftWidth}
\newlength{\algorithmTrialRightWidth}
\newlength{\algorithmTrialHeight}

\begin{figure}[t!]
\begingroup
\setlength{\algorithmTrialLeftWidth}{0.56\textwidth}
\setlength{\algorithmTrialRightWidth}{0.425\textwidth}
\setlength{\algorithmicindent}{0.75em}

\newcommand{\algorithmTrialTightSpacing}{%
\setlength{\parskip}{0pt}%
\setlength{\itemsep}{0pt}%
\setlength{\parsep}{0pt}%
\setlength{\topsep}{0pt}%
\setlength{\partopsep}{0pt}%
\setlength{\abovedisplayskip}{2pt}%
\setlength{\belowdisplayskip}{2pt}%
\setlength{\abovedisplayshortskip}{0pt}%
\setlength{\belowdisplayshortskip}{2pt}%
}

\newcommand{\algorithmTrialParagraph}[1]{%
{\rightskip=0pt\parfillskip=0pt plus 1fil\relax
#1\par}%
}

\newcommand{\algorithmTrialCriticBody}{%
\raggedright
\algorithmTrialTightSpacing
\begin{algorithmic}[1]
\algorithmTrialTightSpacing
\STATE \textbf{Input:} policy $\pi$, discount factor $\gamma$, step-size $\eta$, projection radius $R$, iterations $K$, \mbox{superstate map $\mathcal G$}.%
\STATE \algorithmTrialParagraph{\textbf{Initialize:} sample initialization $\widehat W_0$ as in Sec.~\ref{Preliminaries}; set $\bar W_0=\widehat W_0$, define projection ball as $\mathcal B:=\{W\!:\!\|W^{(h)}-\widehat W_0^{(h)}\|_F\le R,\forall h\in[d]\}$. Set $H_0=\varnothing$, $B_0=\mathcal G(H_0)$, sample $a_0\sim\pi(\cdot\mid B_0)$.}%
\FOR{$k=0,1,\dots,K-1$}
    \STATE Sample $Z_{k+1}\overset{\mathrm{i.i.d.}}{\sim}\mathrm{Bernoulli}(\gamma)$.
    \IF{$Z_{k+1}=0$}
        \STATE \mbox{$(S_{k+1},H_{k+1},B_{k+1})\!\gets\!(S_0\!\sim\!\mathcal D,\varnothing,\varnothing)$},
        \par\nobreak
        sample $a_{k+1}\sim\pi(\cdot\mid B_0)$, $\widehat W_{k+1}\gets\widehat W_k$.
    \ELSE
        \STATE Execute $a_k$, observe $r_k, o_{k+1}$.
        \STATE $H_{k+1}\!\gets\!H_k\|(a_k,o_{k+1}),\ B_{k+1}\!\gets\!\mathcal G(H_{k+1})$.
        \STATE Sample $a_{k+1}\sim\pi(\cdot\mid B_{k+1})$.
        \STATE $y_k\gets r_k+\gamma\widehat Q(B_{k+1},a_{k+1};\widehat W_k)$.
        \STATE $\delta_k\gets \widehat Q(B_k,a_k;\widehat W_k)-y_k$.
        \STATE $\widetilde W_{k+1}\!\gets\!\widehat W_k\!-\!\eta\delta_k\nabla_W\widehat Q(B_k,a_k;\widehat W_k)$.
        \STATE $\widehat W_{k+1}\gets \mathrm{Proj}_{\mathcal B}(\widetilde W_{k+1})$.
    \ENDIF
    \STATE $\bar W_{k+1}\gets \tfrac{k+1}{k+2}\bar W_k+\tfrac{1}{k+2}\widehat W_{k+1}$.
\ENDFOR
\par\vspace{\fill}
\STATE \textbf{Output:} $\widehat Q^\pi:=\widehat Q(\cdot,\cdot;\bar W_K)$.
\end{algorithmic}%
}

\newcommand{\algorithmTrialPolicyBody}{%
\raggedright
\algorithmTrialTightSpacing
\begin{algorithmic}[1]
\algorithmTrialTightSpacing
\setlength{\baselineskip}{\dimexpr\baselineskip\relax plus 1fill}%
\setlength{\lineskip}{1pt plus 1fill}%
\STATE \algorithmTrialParagraph{\textbf{Input:} discount factor $\gamma$, policy stepsize $\kappa$, critic step-size $\eta$, projection radius $R$, number of policy updates $T$, critic iterations $K$, superstate map $\mathcal G$.}
\STATE \textbf{Initialize:} $\pi_0(\cdot\mid B)$ uniform over $\mathcal A$ for every $B\in\mathsf B$.
\FOR{$t=0,1,\dots,T-1$}
    \STATE \algorithmTrialParagraph{Run Alg.~\ref{alg:neural_superstate_td} under $\pi_t$ for $K$ steps~with an independent critic initialization; obtain}
    \STATE $\widehat Q_t(B,a):=\widehat Q^{\pi_t}(B,a),\ \forall(B,a)$.
    \FOR{each $B\in\mathsf B$}
        \STATE Update the policy as
        \[
        \pi_{t+1}(a\mid B)\!\propto\!\pi_t(a\mid B)e^{\kappa\widehat Q_t(B,a)}
        \]
        with
        \[
        \sum_{a\in\mathcal A}\pi_{t+1}(a\mid B)=1.
        \]
    \ENDFOR
\ENDFOR
\STATE \textbf{Output:} last iterate $\pi_T$.
\end{algorithmic}%
}

\sbox{\algorithmTrialCriticBox}{%
\begin{minipage}[t]{\algorithmTrialLeftWidth}
\algorithmTrialCriticBody
\end{minipage}}
\sbox{\algorithmTrialPolicyBox}{%
\begin{minipage}[t]{\algorithmTrialRightWidth}
\algorithmTrialPolicyBody
\end{minipage}}
\setlength{\algorithmTrialHeight}{\dimexpr\ht\algorithmTrialCriticBox+\dp\algorithmTrialCriticBox\relax}
\ifdim\dimexpr\ht\algorithmTrialPolicyBox+\dp\algorithmTrialPolicyBox\relax>\algorithmTrialHeight
  \setlength{\algorithmTrialHeight}{\dimexpr\ht\algorithmTrialPolicyBox+\dp\algorithmTrialPolicyBox\relax}
\fi

\begin{minipage}[t]{\algorithmTrialLeftWidth}
\vspace{0pt}
\begin{algorithm}[H]
\caption{Policy Evaluation.}
\label{alg:neural_superstate_td}
\begin{minipage}[t][\algorithmTrialHeight][s]{\linewidth}
\algorithmTrialCriticBody
\end{minipage}
\end{algorithm}
\end{minipage}\hfill
\begin{minipage}[t]{\algorithmTrialRightWidth}
\vspace{0pt}
\begin{algorithm}[H]
\caption{Policy Update.}
\label{alg:superstate_policy_opt}
\begin{minipage}[t][\algorithmTrialHeight][s]{\linewidth}
\algorithmTrialPolicyBody
\end{minipage}
\end{algorithm}
\end{minipage}
\endgroup
\vspace{-8pt}
\end{figure}

\subsection{Finite-Time Convergence and Regret Analysis}
We next analyze the finite-time convergence and regret performance of our proposed PMD-based policy optimization algorithm for POMDPs. 
The analysis first evaluates the neural TD critic error under POMDP-generated trajectory samples, and then combines this critic guarantee with PMD to obtain a sublinear regret bound against the optimal POMDP value.

{\bf 1) Neural TD Critic:}
For a fixed policy $\pi$, let $\mu^\pi$ be the stationary distribution of the superstate-action Markov chain under $\pi$ with the full-reset mechanism ($\mu^\pi$ is also the normalized discounted visitation distribution of the original SMDP, see Appendix~\ref{app:occupancy-stationarity} for proof).
For critic evaluation, define the weighted $L^2$ norm of $f:\mathsf B\times\mathcal A\to\mathbb R$ by $\|f\|_{\mu^\pi}:=\left(\sum_{B,a}\mu^\pi(B,a)f(B,a)^2\right)^{1/2}$.
To analyze TD learning of our neural critic, we compare the POMDP-generated superstate process with the superstate-action chain whose stationary distribution is $\mu^\pi$.
Toward this end, we derive the following geometric mixing property:

\begingroup
\newtheoremstyle{mixingreview}{\topsep}{\topsep}{\itshape}{}{\bfseries}{.}{.5em}%
{\thmname{#1}\thmnumber{ #2}\thmnote{ (#3)}}
\theoremstyle{mixingreview}
\newtheorem{mixingreviewlemma}[lemma]{Lemma}
\begin{mixingreviewlemma}[Geometric Mixing]
\label{lem:full-reset-mixing}
For a given policy \(\pi\), let \(\{(\tilde B_k,\tilde a_k)\}_{k\ge 0}\) be the superstate-action Markov chain
generated by the superstate MDP with full resets under \(\pi\), which
admits a unique stationary distribution \(\mu^\pi\).
Moreover,
the following bound holds for all \(k\ge 0\),
\[
\sup_{(B,a)\in\mathsf B\times\mathcal A}
\left\|
\mathbb P\bigl((\tilde B_k,\tilde a_k)\in\cdot \mid (\tilde B_0,\tilde a_0)=(B,a)\bigr)-\mu^\pi
\right\|_{\mathrm{TV}}
\le \gamma^k.
\]
\end{mixingreviewlemma}
\endgroup

The geometric mixing property in Lemma~\ref{lem:full-reset-mixing} is commonly imposed as an assumption
in the literature for finite-time TD analyses under Markovian sampling rather than i.i.d. sampling (see, e.g., \citealp{li2020sample,gaur2024closing}).
Notice that we no longer impose this property as an assumption. Rather, we show that the full-reset sampling yields the explicit mixing coefficients.
Intuitively, it measures how fast the superstate-action pair mixes into the stationary distribution from any initial distribution.

We begin with the neural approximation component of the critic analysis. 
Lemma~\ref{lem:approximation-error} shows that, when the projection radius $R$ is sufficiently large, the neural critic can approximate $\tilde Q^\pi$ with a decreasing error as the neural network width and depth increase.

\begin{lemma}[Universal Approximation]
\label{lem:approximation-error}
Consider the policies $\{\pi_t\}_{t=0}^{T-1}$ generated by Algorithm~\ref{alg:superstate_policy_opt}, following from the universal approximation theorem in \citet{shen2022optimal},
there exists a parameter $W_t^\dagger\in\mathbb R^{dm^2}$ for each $0\le t<T$ such that $\bigl\|\widehat Q(\cdot,\cdot;W_t^\dagger)-\tilde Q^{\pi_t}\bigr\|_{\mu^{\pi_t}}
\le
\varepsilon_{\mathrm{app}}(m,d)$,
where $\varepsilon_{\mathrm{app}}(m,d):=\widetilde O\!\left(m^{-2/p}d^{-2/p}\right)$.
Let $\widehat W_{0,t}$ denote the critic initialization at iteration $t$.
Define $R_t^\dagger:=\max_{h\in[d]}\left\|W_t^{\dagger,(h)}-\widehat W_{0,t}^{(h)}\right\|_F$ and $R^\dagger:=\max_{0\le t<T}R_t^\dagger$. Let $\mathcal B_t(R)$ be projection ball of radius $R$ centered at $\widehat W_{0,t}$.
Then, for any projection radius $R\ge R^\dagger$, we have $W_t^\dagger\in\mathcal B_t(R)$ for every $0\le t<T$, and hence
$\max_{0\le t<T}\inf_{W\in\mathcal B_t(R)}\left\|\widehat Q(\cdot,\cdot;W)-\tilde Q^{\pi_t}\right\|_{\mu^{\pi_t}}\le\varepsilon_{\mathrm{app}}(m,d)$.
\end{lemma}

We now turn to the finite-time TD dynamics and establish the following critic error bound.

\begin{theorem}[Critic Error Bound]
\label{thm:critic-error}
Under Assumptions~\ref{assump:bounded-reward} and~\ref{assump:filter-stability},
for a given policy $\pi:=\pi(\cdot | B)$, let $\widehat Q(\cdot,\cdot;\bar W_K)$ be the Q-function learned by Algorithm~\ref{alg:neural_superstate_td}. Suppose the projection radius satisfies $R^\dagger\le R= \widetilde O (m^{1/2}d^{-6})$, where $R^\dagger$ is defined in Lemma~\ref{lem:approximation-error}. 
Then, with probability at least $1-e^{-\Omega(\log^2 m)}$,
\[
\begin{aligned}
&\mathbb E\|\widehat Q(\cdot,\cdot;\bar W_K)-\tilde Q^\pi\|_{\mu^\pi}^2
\leq
\underbrace{\widetilde O\!\left(R^{\frac{8}{3}}m^{-\frac{1}{3}}d^6
+R^{\frac{4}{3}}m^{-\frac{1}{6}}d^3
+R^{\frac{7}{3}}m^{-\frac{1}{6}}d^{\frac{9}{2}}\right)}_{\varepsilon_{\rm NTK}^2}
+\underbrace{\widetilde O\!\left(m^{-\frac{4}{p}}d^{-\frac{4}{p}}\right)}_{\varepsilon_{\rm app}^2}\\
&\quad+
\underbrace{\widetilde O\!\left(\big(d^2+d^{\frac{7}{2}}R+d^5R^2\big)/{\sqrt K}\right)}_{\varepsilon_{\rm TD}^2}
+\underbrace{\widetilde O\!\left((d^{\frac{3}{2}}R+d^3R^2)(1-\rho)^\ell\right)}_{\varepsilon_{\rm hist}^2}
+\underbrace{\widetilde O\!\left((d^{\frac{3}{2}}R+d^3R^2)/{\sqrt K}\right)}_{\varepsilon_{\rm mix}^2}.
\end{aligned}
\]

\end{theorem}

The critic error bound in Theorem~\ref{thm:critic-error} consists of three types of errors, which are interpreted as follows:
\begin{itemize}[leftmargin=*, topsep=2pt, itemsep=2pt, parsep=0pt]
\item 
\textbf{Architecture Error ($\varepsilon_{\mathrm{NTK}}^2$ and $\varepsilon_{\mathrm{app}}^2$):} 
These two terms come from the neural critic architecture.  $\varepsilon_{\mathrm{NTK}}^2$ measures the gap between the nonlinear neural critic and its NTK linearization around the initialization $\widehat W_0$. This term comes from two uses of the NTK approximation in our proof:
1) We compare the learned critic $\widehat Q(\cdot,\cdot;\bar W_K)$ with its linearized version; and 2) During the TD analysis, we replace the nonlinear network values and gradients in the TD update by their linearized counterparts.  $\varepsilon_{\mathrm{NTK}}^2$ is controlled jointly by the width $m$, depth $d$, and projection radius $R$. 
$\varepsilon_{\mathrm{app}}^2$ is the universal-approximation error, measuring how well the neural function class approximates $\tilde Q^\pi$ and decreasing as the width $m$ and depth $d$ of the neural network increase.

\item 
\textbf{Sampling Error ($\varepsilon_{\mathrm{TD}}^2$ and $\varepsilon_{\mathrm{mix}}^2$).}
$\varepsilon_{\mathrm{TD}}^2$ is the standard projected-TD optimization error and decays with the increase of the critic update horizon $K$.
$\varepsilon_{\mathrm{mix}}^2$ is the price we pay for using a single on-policy trajectory instead of an i.i.d. sampling oracle. Under geometric mixing (Lemma~\ref{lem:full-reset-mixing}), choosing the analysis lag $\tau=\max\{1,\lceil\log K/(2|\log\gamma|)\rceil\}$ gives $\gamma^\tau\leq K^{-1/2}$. Hence, this term decays as $\widetilde O(K^{-1/2})$ with the other parameters fixed.

\item 
\textbf{Partial-Observability Error ($\varepsilon_{\mathrm{hist}}^2$).}
$\varepsilon_{\mathrm{hist}}^2$ is unique to PORL and reflects the \emph{model mismatch}: the critic is updated by data from POMDP transitions $(H_k,a_k,r_k,H_{k+1},a_{k+1})$, whereas its Bellman target is defined on the superstate MDP. Under filter stability (Assumption~\ref{assump:filter-stability}), the resulting bias decays as $(1-\rho)^\ell$. Thus increasing the memory length $\ell$ reduces this error exponentially.
\end{itemize}

We note that the critic bound is not monotone in either the width $m$ or the depth $d$ alone, because it also depends on the projection radius $R$. 
The NTK local-linearization argument requires
$
R=\widetilde O(m^{1/2}d^{-6}),
$
whereas Lemma~\ref{lem:approximation-error} requires the projection ball to contain the approximating parameter, i.e., $R^\dagger\le R$. These conditions are compatible when
$
m\gtrsim (R^\dagger)^2d^{12}.
$
Further, if $R^\dagger=O(1)$, then $m\gtrsim d^{12}$ suffices and $d=O(\log m)$ is admissible, yielding a {\em simplified} critic error bound as follows:

\begin{corollary}[Simplified Critic Error Bound]
\label{cor:simplified-critic-bound}
Suppose the conditions of Theorem~\ref{thm:critic-error} hold and assume $R^\dagger=O(1)$.
Choose a projection radius $R=O(1)$ such that $R^\dagger\le R$, and let $d=O(\log m)$. Then, with probability at least $1-e^{-\Omega(\log^2 m)}$, it holds that
${\mathbb E\|\widehat Q^\pi-\tilde Q^\pi\|_{\mu^\pi}^2}
\leq\widetilde O\!\bigl(m^{-1/6}+m^{-4/p}+K^{-1/2}+(1-\rho)^\ell\bigr)$.

\end{corollary}

Corollary~\ref{cor:simplified-critic-bound} implies that, for any $\varepsilon>0$, the critic error can be made at most $\varepsilon$ by taking $m\gtrsim \varepsilon^{-\max\{6,p/4\}}$, $K\gtrsim \varepsilon^{-2}$, and $\ell\gtrsim \log(1/\varepsilon)/|\log(1-\rho)|$. Thus, the neural TD critic can approximate the superstate Bellman fixed point to arbitrary accuracy.

{\bf 2) Policy Update:}
We now combine the neural critic with the PMD policy update. 
Since the critic is based on the sampling distribution $\mu^{\pi_t}$ of the current policy, while the policy update analysis uses the optimal superstate policy $\widetilde\pi^*$ as its comparator, we adopt the following assumption to characterize the distribution mismatch.

\begin{assumption}[Concentrability Coefficient]
\label{assum:coverage}
There exists $\bar{C}_\infty < \infty$ such that, for all $t \ge 0$,
$C_t := \mathbb{E}_{(B,a)\sim \mu^{\pi_t}}\!\left[\left(\frac{\mu^{\widetilde\pi^*}(B,a)}{\mu^{\pi_t}(B,a)}\right)^2\middle|\pi_t\right]$
satisfies
$\mathbb{E}[C_t] \;\le\; \bar{C}_\infty$.
\end{assumption}

The concentrability coefficient $C_t$ quantifies how much the sampling distribution $\mu^{\pi_t}$ covers the distribution
$\mu^{\widetilde\pi^*}$. Intuitively, $C_t$ is small only if $\mu^{\pi_t}$ has full support over the regions visited by
$\widetilde\pi^*$. Assumption~\ref{assum:coverage} is the standard condition used in finite-time analyses of NPG and PMD methods \citep{yuan2022linear,alfano2023novel}, and has also been adopted in POMDPs by \citet{cayci2024finite}.

To evaluate the performance of Algorithm~\ref{alg:superstate_policy_opt}, we use a regret notion similar to \citet{anjarlekar2025scalable}. Let $H_0$ be the initial history and $B_0=\mathcal G(H_0)$ be the corresponding initial superstate. We define
$R_N
:=
\sum_{t=0}^{T-1}\sum_{k=0}^{K-1}
\mathbb E\bigl[V^{\pi^*}(H_0)-\tilde V^{\pi_t}(B_0)\bigr]
=
\sum_{t=0}^{T-1}K\,
\mathbb E\bigl[V^{\pi^*}(H_0)-\tilde V^{\pi_t}(B_0)\bigr]$,
where $N=TK$ is the total interaction budget, and \(V^{\pi^*}(H_0)\) and \(\tilde V^{\pi_t}(B_0)\) denote the value functions of the optimal policy \(\pi^*\) in the original POMDP and the iterate \(\pi_t\) in the superstate MDP, respectively.

\begin{theorem}[Regret Bound]
\label{thm:regret}
Under Assumptions~\ref{assump:bounded-reward}--\ref{assum:coverage},
let $\{\pi_t\}_{t=0}^{T}$ be the policies generated by Algorithm~\ref{alg:superstate_policy_opt},
and suppose that at each policy iteration the critic is trained by
Algorithm~\ref{alg:neural_superstate_td}. Let $N=TK$ denote the total interaction budget, and choose $K=N^{2/3}$ and $T=N^{1/3}$. Then, under the conditions of
Theorem~\ref{thm:critic-error}, with probability at least
$1-N^{1/3}e^{-\Omega(\log^2 m)}$, the total regret satisfies:

\vspace{-.1in}
\begin{equation}
R_N \leq \widetilde O\Bigl(
\underbrace{N(1-\rho)^\ell}_{\mathrm{Model\,\, Approx.}}
\!+\!\underbrace{N(\varepsilon_{\rm NTK}+\varepsilon_{\rm app})}_{\mathrm{Architecture}}
\!+\!\underbrace{N\varepsilon_{\rm hist}}_{\substack{\mathrm{Critic \,\,History}\\ \mathrm{Truncation}}}
\!+\!\underbrace{N^{5/6}P_{\rm mix}}_{\substack{\mathrm{Trajectory}\\ \mathrm{Mixing}}}
\!+\!\underbrace{N^{5/6}P_{\rm opt}}_{\substack{\mathrm{PMD}\\ \mathrm{Optimization}}}
\Bigr),
\vspace{-.1in}
\end{equation}

where $\varepsilon_{\rm NTK},\varepsilon_{\rm app},\varepsilon_{\rm hist}$ are the square roots of the corresponding terms in
Theorem~\ref{thm:critic-error}, and
$P_{\rm mix}(d,R)=d^{3/4}R^{1/2}+d^{3/2}R,
P_{\rm opt}(d,R)=1+d+d^{3/2}R+d^{7/4}R^{1/2}+d^{5/2}R $.

\end{theorem}

The total regret bound in Theorem~\ref{thm:regret} consists of three parts, which can be interpreted as follows:
\begin{itemize}[leftmargin=*, topsep=2pt, itemsep=2pt, parsep=0pt]
\item \textbf{Model Approximation:} By Theorem~\ref{thm:finite-memory-near-optimality}, the term $N(1-\rho)^\ell$ comes from accumulating the value gap between the original POMDP and the surrogate superstate MDP over the total budget $N$, which is controlled by the memory length $\ell$ through filter stability and {\bf decays exponentially fast}.
\item \textbf{Critic-Induced Errors:} 
The middle three terms come from {\em neural} TD evaluation: the architecture term $N(\varepsilon_{\rm NTK}+\varepsilon_{\rm app})$, the critic-level history truncation term $N\varepsilon_{\rm hist}$, and the trajectory-mixing term $N^{5/6}P_{\rm mix}(d,R)$.
The concentrability coefficient in Assumption~\ref{assum:coverage} transfers these critic errors from the sampling distribution $\mu^{\pi_t}$ to the distribution associated with $\widetilde\pi^*$.
As shown in Corollary~\ref{cor:simplified-critic-bound}, the underlying critic errors can be reduced through suitable parameter choices.

\item \textbf{PMD Optimization.} The term $\widetilde O(N^{5/6}P_{\rm opt}(d,R))$ corresponds to the finite-sample optimization errors from both critic evaluation and policy update, which is {\bf sublinear} in $N$.
\end{itemize}

The critic-induced terms have been characterized in Theorem~\ref{thm:critic-error} and Corollary~\ref{cor:simplified-critic-bound}. 
The following corollary offers a clean and simplified selection of hyperparameters under the assumption $R^\dagger=O(1)$: 

\begin{corollary}[Sublinear Regret]
\label{cor:sublinear-regret}
Under the conditions of Corollary~\ref{cor:simplified-critic-bound}, let
$K=N^{2/3}$ and $T=N^{1/3}$.
If the memory length $\ell$ and width $m$ satisfy that $\ell\ge \log N/(3|\log(1-\rho)|)$, and $m\gtrsim \max\{N^2,N^{p/12}\}$,
then, with probability at least $1-N^{1/3}e^{-\Omega(\log^2 m)}$, the regret of
Algorithm~\ref{alg:superstate_policy_opt} satisfies 
$R_N=$
$\widetilde O(N^{5/6})$.

\end{corollary}
Algorithm~\ref{alg:superstate_policy_opt} 
achieves a sublinear regret $\widetilde O(N^{5/6})$.
The universal approximation property of the neural critic, combined with a sufficiently large width $m$ and logarithmic memory length $\ell$, ensures the overall regret remains {\em sublinear} in $N$.

\section{Experimental Results}

\begin{wrapfigure}[8]{r}{0.28\linewidth}
    \centering
    \vspace{-12pt}
    \includegraphics[width=\linewidth]{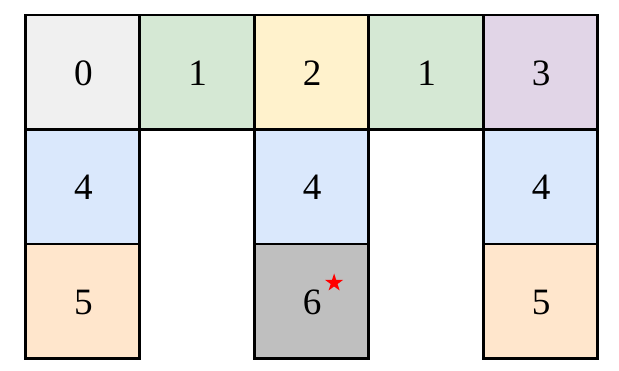}
    \vspace{-.15in}
    \caption{CheeseMaze.}
    \label{fig:cheesemaze}
    \vspace{-.1in}
\end{wrapfigure}
We evaluate our policy optimization method on the \textit{CheeseMaze} benchmark shown in Fig.~\ref{fig:cheesemaze}, a partially observable navigation task introduced in \citet{mccallum1993overcoming}. 
The environment contains 11 latent states and 7 observations, where multiple states share the same observation.
The action space consists of 4 directions.
The agent should navigate to the goal state with observation 6, and receives a reward of \(+1\) only upon reaching the goal and \(0\) otherwise.
We compare against two standard PORL baselines: Q-learning and Actor-Critic.
All methods use the same finite-memory superstate input and the same interaction budget. 

\begin{wrapfigure}[14]{R}{0.4\textwidth}
\vspace{-.0in}
\centering
\includegraphics[width=0.38\textwidth]{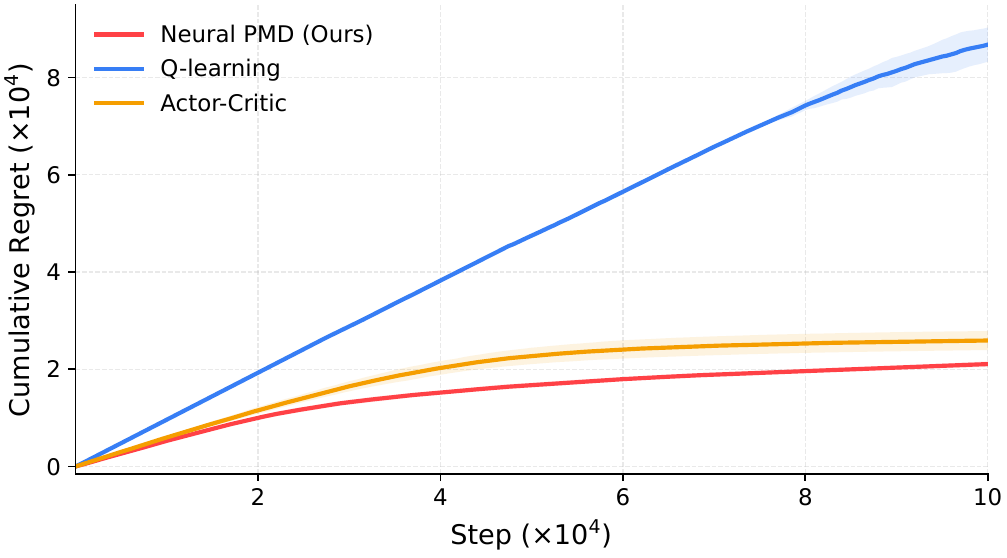}
\vspace{-.1in}
\caption{Empirical regret (\(\ell\!=\!2\)):
The y-axis reports the empirical cumulative regret $\widehat R_N:=\sum_{t=0}^{T-1}K(V^*-\widehat V_t)$.
Curves show mean $\pm 1$ standard deviation.
}
\label{fig:empirical_regret}
\vspace{-.2in}
\end{wrapfigure}
\begin{figure*}[t]
\centering
\includegraphics[width=0.9\textwidth]{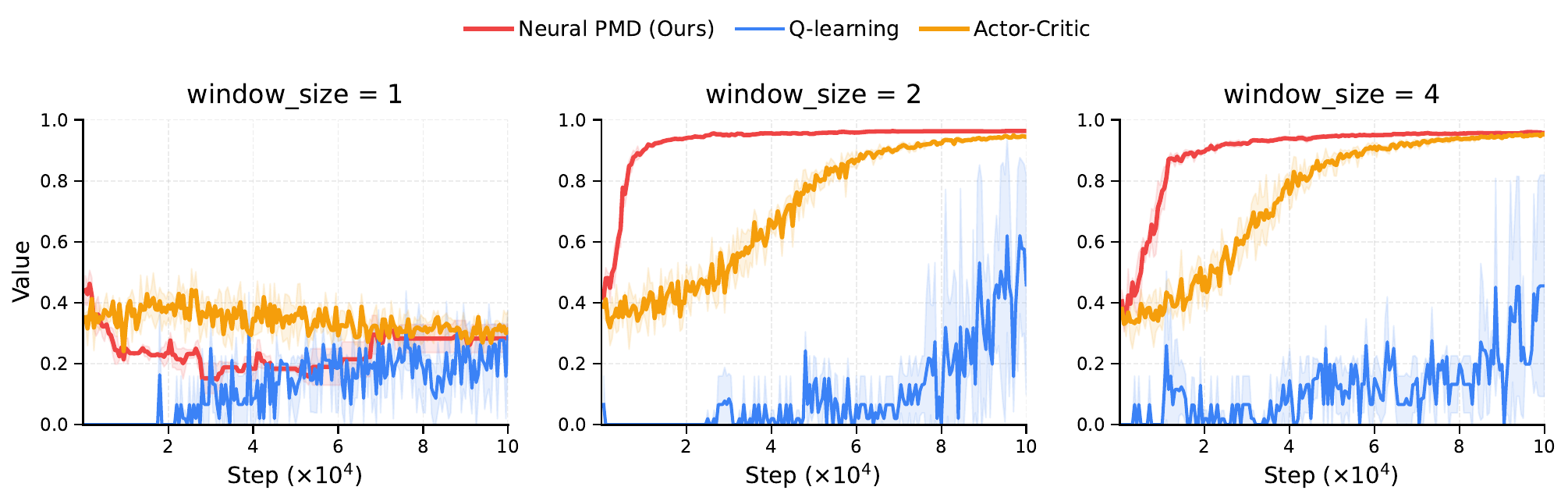}
\vspace{-.15in}
\caption{Effect of superstate window size on policy value.
The y-axis reports value $\widehat V=\mathbb E[\sum_h \gamma^h r_h]$ estimated over 50 episodes every 500 training steps.
Curves show mean $\pm$ one standard deviation. %
}
\label{fig:window_size_cheesemaze}
\vspace{-.2in}
\end{figure*}

Fig.~\ref{fig:window_size_cheesemaze} compares our method with Q-learning and Actor-Critic under different superstate window sizes $\ell\in\{1,2,4\}$. 
When $\ell=1$, the agent only observes the most recent action--observation pair, giving a minimal-memory baseline. All methods perform poorly in this setting.
As the window size increases to $\ell=2$ and $\ell=4$, our method becomes more stable than Q-learning and converges earlier than Actor-Critic. 
Q-learning is unstable because its greedy max-based update can amplify estimation errors. 
Actor-Critic improves more smoothly, but is slower because it also learns actor parameters. 
Increasing $\ell$ from $2$ to $4$ yields only marginal improvement, suggesting that a short finite-memory window already captures most useful information in this benchmark.
Fig.~\ref{fig:empirical_regret} shows that the regret curve of our neural PMD-based method grows more slowly and stays below the baseline curves, which indicates a sublinear empirical regret consistent with our analysis.
Sensitivity to policy stepsize $\kappa$
and additional hyper-parameters are deferred to the Appendix~\ref{app:experiments}.

\section{Conclusion}
In this paper, we studied RL in POMDPs with neural function approximation through a finite-memory superstate MDP.
We first characterized the superstate MDP approximation, and then proved a finite-time critic error bound and a sublinear regret bound against the optimal POMDP value.
We conducted experiments on the CheeseMaze benchmark to validate our theoretical results, showing greater stability than Q-learning and higher sample efficiency than Actor–Critic.
A limitation of this work is that it focuses on finite state, observation, and action spaces, which allows truncated histories to enable a finite superstate MDP.
Future directions include sharpening theoretical results, relaxing structural assumptions, 
extending the framework to more sophisticated architectures (e.g., Transformers), and generalizing it to continuous state and action spaces.

\bibliography{neurips2026_reference}

\begin{thebibliography}{51}
\providecommand{\natexlab}[1]{#1}
\providecommand{\url}[1]{\texttt{#1}}
\expandafter\ifx\csname urlstyle\endcsname\relax
  \providecommand{\doi}[1]{doi: #1}\else
  \providecommand{\doi}{doi: \begingroup \urlstyle{rm}\Url}\fi

\bibitem[Abbasi-Yadkori et~al.(2019)Abbasi-Yadkori, Bartlett, Bhatia, Lazic,
  Szepesvari, and Weisz]{abbasi2019politex}
Yasin Abbasi-Yadkori, Peter Bartlett, Kush Bhatia, Nevena Lazic, Csaba
  Szepesvari, and Gell{\'e}rt Weisz.
\newblock Politex: Regret bounds for policy iteration using expert prediction.
\newblock In \emph{International conference on machine learning}, pp.\
  3692--3702. PMLR, 2019.

\bibitem[Agarwal et~al.(2021)Agarwal, Kakade, Lee, and
  Mahajan]{agarwal2021theory}
Alekh Agarwal, Sham~M Kakade, Jason~D Lee, and Gaurav Mahajan.
\newblock On the theory of policy gradient methods: Optimality, approximation,
  and distribution shift.
\newblock \emph{Journal of Machine Learning Research}, 22\penalty0
  (98):\penalty0 1--76, 2021.

\bibitem[Alfano et~al.(2023)Alfano, Yuan, and Rebeschini]{alfano2023novel}
Carlo Alfano, Rui Yuan, and Patrick Rebeschini.
\newblock A novel framework for policy mirror descent with general
  parameterization and linear convergence.
\newblock \emph{Advances in Neural Information Processing Systems},
  36:\penalty0 30681--30725, 2023.

\bibitem[Amato et~al.(2012)Amato, Bernstein, and
  Zilberstein]{amato2012optimizing}
Christopher Amato, Daniel~S Bernstein, and Shlomo Zilberstein.
\newblock Optimizing memory-bounded controllers for decentralized pomdps.
\newblock \emph{arXiv preprint arXiv:1206.5258}, 2012.

\bibitem[Anjarlekar et~al.(2025)Anjarlekar, Etesami, and
  Srikant]{anjarlekar2025scalable}
Ameya Anjarlekar, Rasoul Etesami, and R~Srikant.
\newblock Scalable policy-based rl algorithms for pomdps.
\newblock \emph{arXiv preprint arXiv:2510.06540}, 2025.

\bibitem[Bhandari et~al.(2018)Bhandari, Russo, and Singal]{bhandari2018finite}
Jalaj Bhandari, Daniel Russo, and Raghav Singal.
\newblock A finite time analysis of temporal difference learning with linear
  function approximation.
\newblock In \emph{Conference on learning theory}, pp.\  1691--1692. PMLR,
  2018.

\bibitem[Cai et~al.(2019)Cai, Yang, Lee, and Wang]{cai2019neural}
Q~Cai, Zhuoran Yang, Jason Lee, and Zhaoran Wang.
\newblock Neural temporal-difference and q-learning provably converge to global
  optima. arxiv.
\newblock \emph{Learning}, 2019.

\bibitem[Cai et~al.(2024)Cai, Liu, Oikonomou, and Zhang]{cai2024provable}
Yang Cai, Xiangyu Liu, Argyris Oikonomou, and Kaiqing Zhang.
\newblock Provable partially observable reinforcement learning with privileged
  information.
\newblock \emph{Advances in Neural Information Processing Systems},
  37:\penalty0 63790--63857, 2024.

\bibitem[Cayci \& Eryilmaz(2024)Cayci and Eryilmaz]{cayci2024recurrent}
Semih Cayci and Atilla Eryilmaz.
\newblock Recurrent natural policy gradient for pomdps.
\newblock \emph{arXiv preprint arXiv:2405.18221}, 2024.

\bibitem[Cayci et~al.(2024)Cayci, He, and Srikant]{cayci2024finite}
Semih Cayci, Niao He, and R~Srikant.
\newblock Finite-time analysis of natural actor-critic for pomdps.
\newblock \emph{SIAM Journal on Mathematics of Data Science}, 6\penalty0
  (4):\penalty0 869--896, 2024.

\bibitem[Cen et~al.(2022)Cen, Cheng, Chen, Wei, and Chi]{cen2022fast}
Shicong Cen, Chen Cheng, Yuxin Chen, Yuting Wei, and Yuejie Chi.
\newblock Fast global convergence of natural policy gradient methods with
  entropy regularization.
\newblock \emph{Operations Research}, 70\penalty0 (4):\penalty0 2563--2578,
  2022.

\bibitem[Debrabant \& Stannat(2013)Debrabant and
  Stannat]{debrabant2013stability}
Birgit Debrabant and Wilhelm Stannat.
\newblock Stability of the optimal filter in a hidden markov model with
  multiplicative noise.
\newblock \emph{arXiv preprint arXiv:1301.4428}, 2013.

\bibitem[Gaur et~al.(2024)Gaur, Bedi, Wang, and Aggarwal]{gaur2024closing}
Mudit Gaur, Amrit~Singh Bedi, Di~Wang, and Vaneet Aggarwal.
\newblock Closing the gap: Achieving global convergence (last iterate) of
  actor-critic under markovian sampling with neural network parametrization.
\newblock \emph{arXiv preprint arXiv:2405.01843}, 2024.

\bibitem[Golowich et~al.(2023)Golowich, Moitra, and
  Rohatgi]{golowich2023planning}
Noah Golowich, Ankur Moitra, and Dhruv Rohatgi.
\newblock Planning and learning in partially observable systems via filter
  stability.
\newblock In \emph{Proceedings of the 55th Annual ACM Symposium on Theory of
  Computing}, pp.\  349--362, 2023.

\bibitem[Hausknecht \& Stone(2015)Hausknecht and Stone]{hausknecht2015deep}
Matthew~J Hausknecht and Peter Stone.
\newblock Deep recurrent q-learning for partially observable mdps.
\newblock In \emph{AAAI fall symposia}, volume~45, pp.\  141, 2015.

\bibitem[Hauskrecht \& Fraser(2000)Hauskrecht and
  Fraser]{hauskrecht2000planning}
Milos Hauskrecht and Hamish Fraser.
\newblock Planning treatment of ischemic heart disease with partially
  observable markov decision processes.
\newblock \emph{Artificial intelligence in medicine}, 18\penalty0 (3):\penalty0
  221--244, 2000.

\bibitem[Hornik et~al.(1989)Hornik, Stinchcombe, and
  White]{hornik1989multilayer}
Kurt Hornik, Maxwell Stinchcombe, and Halbert White.
\newblock Multilayer feedforward networks are universal approximators.
\newblock \emph{Neural networks}, 2\penalty0 (5):\penalty0 359--366, 1989.

\bibitem[Huang et~al.(2022)Huang, Dossa, Ye, Braga, Chakraborty, Mehta, and
  Ara{\~A}{\v{s}}jo]{huang2022cleanrl}
Shengyi Huang, Rousslan Fernand~Julien Dossa, Chang Ye, Jeff Braga, Dipam
  Chakraborty, Kinal Mehta, and Jo{\~A}{\c{G}}o~GM Ara{\~A}{\v{s}}jo.
\newblock Cleanrl: High-quality single-file implementations of deep
  reinforcement learning algorithms.
\newblock \emph{Journal of Machine Learning Research}, 23\penalty0
  (274):\penalty0 1--18, 2022.

\bibitem[Igl et~al.(2018)Igl, Zintgraf, Le, Wood, and Whiteson]{igl2018deep}
Maximilian Igl, Luisa Zintgraf, Tuan~Anh Le, Frank Wood, and Shimon Whiteson.
\newblock Deep variational reinforcement learning for pomdps.
\newblock In \emph{International conference on machine learning}, pp.\
  2117--2126. PMLR, 2018.

\bibitem[Kaelbling et~al.(1998)Kaelbling, Littman, and
  Cassandra]{kaelbling1998planning}
Leslie~Pack Kaelbling, Michael~L Littman, and Anthony~R Cassandra.
\newblock Planning and acting in partially observable stochastic domains.
\newblock \emph{Artificial intelligence}, 101\penalty0 (1-2):\penalty0 99--134,
  1998.

\bibitem[Kakade(2001)]{kakade2001natural}
Sham~M Kakade.
\newblock A natural policy gradient.
\newblock \emph{Advances in neural information processing systems}, 14, 2001.

\bibitem[Kara \& Yuksel(2022)Kara and Yuksel]{kara2022near}
Ali Kara and Serdar Yuksel.
\newblock Near optimality of finite memory feedback policies in partially
  observed markov decision processes.
\newblock \emph{Journal of Machine Learning Research}, 23\penalty0
  (11):\penalty0 1--46, 2022.

\bibitem[Kara \& Y{\"u}ksel(2023)Kara and Y{\"u}ksel]{kara2023convergence}
Ali~Devran Kara and Serdar Y{\"u}ksel.
\newblock Convergence of finite memory q learning for pomdps and near
  optimality of learned policies under filter stability.
\newblock \emph{Mathematics of Operations Research}, 48\penalty0 (4):\penalty0
  2066--2093, 2023.

\bibitem[Kurniawati(2021)]{kurniawati2021partially}
Hanna Kurniawati.
\newblock Partially observable markov decision processes (pomdps) and robotics.
\newblock \emph{arXiv preprint arXiv:2107.07599}, 2021.

\bibitem[Lan(2023)]{lan2023policy}
Guanghui Lan.
\newblock Policy mirror descent for reinforcement learning: Linear convergence,
  new sampling complexity, and generalized problem classes.
\newblock \emph{Mathematical programming}, 198\penalty0 (1):\penalty0
  1059--1106, 2023.

\bibitem[Li et~al.(2020)Li, Wei, Chi, Gu, and Chen]{li2020sample}
Gen Li, Yuting Wei, Yuejie Chi, Yuantao Gu, and Yuxin Chen.
\newblock Sample complexity of asynchronous q-learning: Sharper analysis and
  variance reduction.
\newblock \emph{Advances in neural information processing systems},
  33:\penalty0 7031--7043, 2020.

\bibitem[Louren{\c{c}}o et~al.(2018)Louren{\c{c}}o, Batista, Oliveira, and
  Silvestre]{lourenco2018globally}
Pedro Louren{\c{c}}o, Pedro Batista, Paulo Oliveira, and Carlos Silvestre.
\newblock A globally exponentially stable filter for bearing-only simultaneous
  localization and mapping with monocular vision.
\newblock \emph{Robotics and Autonomous Systems}, 100:\penalty0 61--77, 2018.

\bibitem[Lovejoy(1991)]{lovejoy1991survey}
William~S Lovejoy.
\newblock A survey of algorithmic methods for partially observed markov
  decision processes.
\newblock \emph{Annals of Operations Research}, 28\penalty0 (1):\penalty0
  47--65, 1991.

\bibitem[McCallum(1993)]{mccallum1993overcoming}
Andrew McCallum.
\newblock Overcoming incomplete perception with util distinction memory.
\newblock In \emph{Proceedings of the Tenth International Conference on
  International Conference on Machine Learning}, pp.\  190--196, 1993.

\bibitem[McDonald \& Y{\"u}ksel(2020)McDonald and
  Y{\"u}ksel]{mcdonald2020exponential}
Curtis McDonald and Serdar Y{\"u}ksel.
\newblock Exponential filter stability via dobrushin’s coefficient.
\newblock 2020.

\bibitem[Mei et~al.(2020)Mei, Xiao, Szepesvari, and Schuurmans]{mei2020global}
Jincheng Mei, Chenjun Xiao, Csaba Szepesvari, and Dale Schuurmans.
\newblock On the global convergence rates of softmax policy gradient methods.
\newblock In \emph{International conference on machine learning}, pp.\
  6820--6829. PMLR, 2020.

\bibitem[Meuleau et~al.(2013)Meuleau, Peshkin, Kim, and
  Kaelbling]{meuleau2013learning}
Nicolas Meuleau, Leonid Peshkin, Kee-Eung Kim, and Leslie~Pack Kaelbling.
\newblock Learning finite-state controllers for partially observable
  environments.
\newblock \emph{arXiv preprint arXiv:1301.6721}, 2013.

\bibitem[Morad et~al.(2023)Morad, Kortvelesy, Bettini, Liwicki, and
  Prorok]{morad2023popgym}
Steven Morad, Ryan Kortvelesy, Matteo Bettini, Stephan Liwicki, and Amanda
  Prorok.
\newblock Popgym: Benchmarking partially observable reinforcement learning.
\newblock \emph{arXiv preprint arXiv:2303.01859}, 2023.

\bibitem[Pascanu et~al.(2013)Pascanu, Mikolov, and
  Bengio]{pascanu2013difficulty}
Razvan Pascanu, Tomas Mikolov, and Yoshua Bengio.
\newblock On the difficulty of training recurrent neural networks.
\newblock In \emph{International conference on machine learning}, pp.\
  1310--1318. Pmlr, 2013.

\bibitem[Poupart \& Boutilier(2003)Poupart and Boutilier]{poupart2003bounded}
Pascal Poupart and Craig Boutilier.
\newblock Bounded finite state controllers.
\newblock \emph{Advances in neural information processing systems}, 16, 2003.

\bibitem[Schulman et~al.(2015)Schulman, Levine, Abbeel, Jordan, and
  Moritz]{schulman2015trust}
John Schulman, Sergey Levine, Pieter Abbeel, Michael Jordan, and Philipp
  Moritz.
\newblock Trust region policy optimization.
\newblock In \emph{International conference on machine learning}, pp.\
  1889--1897. PMLR, 2015.

\bibitem[Schulman et~al.(2017)Schulman, Wolski, Dhariwal, Radford, and
  Klimov]{schulman2017proximal}
John Schulman, Filip Wolski, Prafulla Dhariwal, Alec Radford, and Oleg Klimov.
\newblock Proximal policy optimization algorithms.
\newblock \emph{arXiv preprint arXiv:1707.06347}, 2017.

\bibitem[Shani et~al.(2020)Shani, Efroni, and Mannor]{shani2020adaptive}
Lior Shani, Yonathan Efroni, and Shie Mannor.
\newblock Adaptive trust region policy optimization: Global convergence and
  faster rates for regularized mdps.
\newblock In \emph{Proceedings of the AAAI conference on artificial
  intelligence}, volume~34, pp.\  5668--5675, 2020.

\bibitem[Shen et~al.(2022)Shen, Yang, and Zhang]{shen2022optimal}
Zuowei Shen, Haizhao Yang, and Shijun Zhang.
\newblock Optimal approximation rate of relu networks in terms of width and
  depth.
\newblock \emph{Journal de Math{\'e}matiques Pures et Appliqu{\'e}es},
  157:\penalty0 101--135, 2022.

\bibitem[Smallwood \& Sondik(1973)Smallwood and Sondik]{smallwood1973optimal}
Richard~D Smallwood and Edward~J Sondik.
\newblock The optimal control of partially observable markov processes over a
  finite horizon.
\newblock \emph{Operations research}, 21\penalty0 (5):\penalty0 1071--1088,
  1973.

\bibitem[Sun et~al.(2022)Sun, Li, and Wang]{sun2022finite}
Tao Sun, Dongsheng Li, and Bao Wang.
\newblock Finite-time analysis of adaptive temporal difference learning with
  deep neural networks.
\newblock \emph{Advances in Neural Information Processing Systems},
  35:\penalty0 19592--19604, 2022.

\bibitem[Sunberg \& Kochenderfer(2022)Sunberg and
  Kochenderfer]{sunberg2022improving}
Zachary Sunberg and Mykel~J Kochenderfer.
\newblock Improving automated driving through pomdp planning with human
  internal states.
\newblock \emph{IEEE Transactions on Intelligent Transportation Systems},
  23\penalty0 (11):\penalty0 20073--20083, 2022.

\bibitem[Tomar et~al.(2020)Tomar, Shani, Efroni, and
  Ghavamzadeh]{tomar2020mirror}
Manan Tomar, Lior Shani, Yonathan Efroni, and Mohammad Ghavamzadeh.
\newblock Mirror descent policy optimization.
\newblock \emph{arXiv preprint arXiv:2005.09814}, 2020.

\bibitem[Whiteley(2021)]{whiteley2021dimension}
Nick Whiteley.
\newblock Dimension-free wasserstein contraction of nonlinear filters.
\newblock \emph{Stochastic Processes and their Applications}, 135:\penalty0
  31--50, 2021.

\bibitem[Wierstra et~al.(2007)Wierstra, Foerster, Peters, and
  Schmidhuber]{wierstra2007solving}
Daan Wierstra, Alexander Foerster, Jan Peters, and Juergen Schmidhuber.
\newblock Solving deep memory pomdps with recurrent policy gradients.
\newblock In \emph{International conference on artificial neural networks},
  pp.\  697--706. Springer, 2007.

\bibitem[Xiao(2022)]{xiao2022convergence}
Lin Xiao.
\newblock On the convergence rates of policy gradient methods.
\newblock \emph{Journal of Machine Learning Research}, 23\penalty0
  (282):\penalty0 1--36, 2022.

\bibitem[Yuan et~al.(2022)Yuan, Du, Gower, Lazaric, and Xiao]{yuan2022linear}
Rui Yuan, Simon~S Du, Robert~M Gower, Alessandro Lazaric, and Lin Xiao.
\newblock Linear convergence of natural policy gradient methods with log-linear
  policies.
\newblock \emph{arXiv preprint arXiv:2210.01400}, 2022.

\bibitem[Zhan et~al.(2023)Zhan, Cen, Huang, Chen, Lee, and Chi]{zhan2023policy}
Wenhao Zhan, Shicong Cen, Baihe Huang, Yuxin Chen, Jason~D Lee, and Yuejie Chi.
\newblock Policy mirror descent for regularized reinforcement learning: A
  generalized framework with linear convergence.
\newblock \emph{SIAM Journal on Optimization}, 33\penalty0 (2):\penalty0
  1061--1091, 2023.

\bibitem[Zhang et~al.(2025{\natexlab{a}})Zhang, Geng, Yu, Yin, Zhang, Tan,
  Zhou, Li, Xue, Li, et~al.]{zhang2025landscape}
Guibin Zhang, Hejia Geng, Xiaohang Yu, Zhenfei Yin, Zaibin Zhang, Zelin Tan,
  Heng Zhou, Zhongzhi Li, Xiangyuan Xue, Yijiang Li, et~al.
\newblock The landscape of agentic reinforcement learning for llms: A survey.
\newblock \emph{arXiv preprint arXiv:2509.02547}, 2025{\natexlab{a}}.

\bibitem[Zhang et~al.(2024)Zhang, Lu, and Zhao]{zhang2024deep}
Shijun Zhang, Jianfeng Lu, and Hongkai Zhao.
\newblock Deep network approximation: Beyond relu to diverse activation
  functions.
\newblock \emph{Journal of Machine Learning Research}, 25\penalty0
  (35):\penalty0 1--39, 2024.

\bibitem[Zhang et~al.(2025{\natexlab{b}})Zhang, Oh, Hairi, Luo, Velasquez, and
  Liu]{zhang2025finite}
Zhiyao Zhang, Myeung~Suk Oh, FNU Hairi, Ziyue Luo, Alvaro Velasquez, and Jia
  Liu.
\newblock Finite-time global optimality convergence in deep neural actor-critic
  methods for decentralized multi-agent reinforcement learning.
\newblock \emph{arXiv preprint arXiv:2505.18433}, 2025{\natexlab{b}}.

\end{thebibliography}
\bibliographystyle{iclr2027_conference} %
\newpage
\appendix
\clearpage

\section*{Appendix}
\section{Proof of POMDP-to-SMDP Approximation}
\label{app:pomdp-smdp-approx}
\subsection{Preliminaries}

\begin{lemma}[Superstate-belief approximation]
\label{lem:pairwise-to-superstate}
Under Assumption~\ref{assump:filter-stability}, for every history $H$ and $B=\mathcal G(H)$,
\[
\|b_H-b_B\|_{\mathrm{TV}}
\le
(1-\rho)^\ell,
\]
where $b_B$ is the fixed Bayes-filter belief defined in Section~\ref{Preliminaries}.
\end{lemma}

\begin{proof}
If $|B|<\ell$, no prefix has been discarded. Thus, $H=B$ and $b_H=b_B$. If $|B|=\ell$, the finite sequence $B$ can itself be considered as an admissible history initialized from $\mathcal D$. Moreover, $\mathcal G(B)=B=\mathcal G(H)$.
Applying Assumption~\ref{assump:filter-stability} directly to the histories $H$ and $B$ gives
\[
\|b_H-b_B\|_{\mathrm{TV}}
\le
(1-\rho)^\ell.
\]
\end{proof}

\begin{lemma}[Reward and transition mismatch under filter stability]
\label{lem:reward-transition-mismatch}
Under Assumptions~\ref{assump:bounded-reward} and~\ref{assump:filter-stability}, for every history-action pair $(H,a)$, letting $B=\mathcal G(H)$, we have
\[
\left|\mathbb E[r_k\mid H_k=H,a_k=a]-\tilde r(B,a)\right|\le 2r_{\max}(1-\rho)^\ell .
\]
Moreover, for any superstate policy $\pi$, if $a_{k+1}\sim\pi(\cdot\mid B_{k+1})$, then

\[
\left\|\mathbb P\bigl((B_{k+1},a_{k+1})\in\cdot\mid H_k=H,a_k=a\bigr)-\widetilde{\mathcal P}^{\pi}(\cdot\mid B,a)\right\|_{\mathrm{TV}}\le (1-\rho)^\ell .
\]

\end{lemma}

\begin{proof}
Fix any history-action pair $(H,a)$ and let $B=\mathcal G(H)$. For the reward mismatch, the POMDP-generated reward satisfies $\mathbb E[r_k\mid H_k=H,a_k=a]=\sum_s b_H(s)r(s,a)$, whereas the superstate reward is $\tilde r(B,a)=\sum_s b_B(s)r(s,a)$. Therefore,
\[
\begin{aligned}
\left|\mathbb E[r_k\mid H_k=H,a_k=a]-\tilde r(B,a)\right|
&=
\left|\sum_s \bigl(b_H(s)-b_B(s)\bigr)r(s,a)\right| \\
&\le r_{\max}\sum_s |b_H(s)-b_B(s)| \\
&=2r_{\max}\|b_H-b_B\|_{\mathrm{TV}} \\
&\le 2r_{\max}(1-\rho)^\ell,
\end{aligned}
\]
where the last inequality follows from Lemma~\ref{lem:pairwise-to-superstate}.

For a fixed policy $\pi$, define the superstate-action transition kernel $\widetilde{\mathcal P}^{\pi}(B',a'\mid B,a):=\widetilde{\mathcal P}(B'\mid B,a)\pi(a'\mid B')$.
For the transition mismatch, recall that the POMDP-generated next-superstate transition and the superstate MDP transition are
\[
\resizebox{\linewidth}{!}{$\displaystyle
\begin{aligned}
\mathbb P(B_{k+1}=B'\mid H_k=H,a_k=a)
&=\sum_{o,s,s'}\mathbf 1\{\mathcal G(H\parallel\{a,o\})=B'\}\Phi(o\mid s')P(s'\mid s,a)b_H(s),\\
\widetilde{\mathcal P}(B'\mid B,a)
&=\sum_{o,s,s'}\mathbf 1\{\mathcal G(B\parallel\{a,o\})=B'\}\Phi(o\mid s')P(s'\mid s,a)b_B(s).
\end{aligned}
$}
\]
Since $B=\mathcal G(H)$, appending $(a,o)$ and then truncating gives $\mathcal G(H\parallel\{a,o\})=\mathcal G(B\parallel\{a,o\})$. Hence the two kernels differ only through the belief terms $b_H$ and $b_B$. After sampling $a_{k+1}\sim\pi(\cdot\mid B_{k+1})$, for every $(B',a')$ we have
\[
\begin{aligned}
&\mathbb P(B_{k+1}=B',a_{k+1}=a'\mid H_k=H,a_k=a)-\widetilde{\mathcal P}^{\pi}(B',a'\mid B,a) \\
&\qquad=
\pi(a'\mid B')\sum_{o,s,s'}\mathbf 1\{\mathcal G(B\parallel\{a,o\})=B'\}\Phi(o\mid s')P(s'\mid s,a)\bigl(b_H(s)-b_B(s)\bigr).
\end{aligned}
\]
Therefore,
\[
\resizebox{\linewidth}{!}{$\displaystyle
\begin{aligned}
&\left\|\mathbb P\bigl((B_{k+1},a_{k+1})\in\cdot\mid H_k=H,a_k=a\bigr)-\widetilde{\mathcal P}^{\pi}(\cdot\mid B,a)\right\|_{\mathrm{TV}} \\
&\qquad=
\frac12\sum_{B',a'}\left|\mathbb P(B_{k+1}=B',a_{k+1}=a'\mid H_k=H,a_k=a)-\widetilde{\mathcal P}^{\pi}(B',a'\mid B,a)\right| \\
&\qquad\le
\frac12\sum_{B',a'}\pi(a'\mid B')\sum_{o,s,s'}\mathbf 1\{\mathcal G(B\parallel\{a,o\})=B'\}\Phi(o\mid s')P(s'\mid s,a)|b_H(s)-b_B(s)| \\
&\qquad=
\frac12\sum_s |b_H(s)-b_B(s)|\sum_{o,s'}\Phi(o\mid s')P(s'\mid s,a)\sum_{B'}\mathbf 1\{\mathcal G(B\parallel\{a,o\})=B'\}\sum_{a'}\pi(a'\mid B') \\
&\qquad=
\frac12\sum_s |b_H(s)-b_B(s)| \\
&\qquad=
\|b_H-b_B\|_{\mathrm{TV}}\\
&\qquad \le
(1-\rho)^\ell,
\end{aligned}
$}
\]
where the last inequality follows from Lemma~\ref{lem:pairwise-to-superstate}.
\end{proof}

\subsection{Proof of Theorem~\ref{thm:finite-memory-near-optimality}}
\label{app:optimality-bridge}
\begin{proof}
Let $\pi^*$ be an unrestricted optimal policy of the original POMDP. Let $\widetilde\pi^*$ be an optimal policy of the induced superstate MDP. Define $\Delta:=\sup_{H\in\mathbb H}|V^{\pi^*}(H)-\tilde V^{\widetilde\pi^*}(\mathcal G(H))|$. For any history $H$ with $B=\mathcal G(H)$, the Bellman optimality equations give
\[
\begin{aligned}
&\bigl|V^{\pi^*}(H)-\tilde V^{\widetilde\pi^*}(B)\bigr|
\le \max_{a\in\mathcal A}\Bigl\{\bigl|\mathbb E[r_k\mid H_k=H,a_k=a]-\tilde r(B,a)\bigr|\\
&\qquad+\gamma\bigl|\mathbb E[V^{\pi^*}(H_{k+1})\mid H_k=H,a_k=a]
-\mathbb E_{\widetilde{\mathcal P}(\cdot\mid B,a)}[\tilde V^{\widetilde\pi^*}(B_{k+1})]\bigr|\Bigr\}.
\end{aligned}
\]
The inequality uses $|\max_a f_a-\max_a g_a|\le\max_a|f_a-g_a|$ and the triangle inequality. For each action $a$, we add and subtract the conditional expectation $\mathbb E[\tilde V^{\widetilde\pi^*}(B_{k+1})\mid H_k=H,a_k=a]$ under the original POMDP to obtain
\[
\begin{aligned}
&\mathbb E[V^{\pi^*}(H_{k+1})\mid H_k=H,a_k=a]
-\mathbb E_{\widetilde{\mathcal P}(\cdot\mid B,a)}[\tilde V^{\widetilde\pi^*}(B_{k+1})]\\
&=\mathbb E[V^{\pi^*}(H_{k+1})\mid H_k=H,a_k=a]
-\mathbb E[\tilde V^{\widetilde\pi^*}(B_{k+1})\mid H_k=H,a_k=a]\\
&\quad+\mathbb E[\tilde V^{\widetilde\pi^*}(B_{k+1})\mid H_k=H,a_k=a]
-\mathbb E_{\widetilde{\mathcal P}(\cdot\mid B,a)}[\tilde V^{\widetilde\pi^*}(B_{k+1})].
\end{aligned}
\]
The triangle inequality and $|\mathbb E[X]|\le\mathbb E[|X|]$ therefore imply
\[
\begin{aligned}
&\bigl|V^{\pi^*}(H)-\tilde V^{\widetilde\pi^*}(B)\bigr|
\le\max_{a\in\mathcal A}\Bigl\{\bigl|\mathbb E[r_k\mid H_k=H,a_k=a]-\tilde r(B,a)\bigr|\\
&\qquad+\gamma\mathbb E\!\left[\bigl|V^{\pi^*}(H_{k+1})-\tilde V^{\widetilde\pi^*}(B_{k+1})\bigr|\mid H_k=H,a_k=a\right]\\
&\qquad+\gamma\bigl|\mathbb E[\tilde V^{\widetilde\pi^*}(B_{k+1})\mid H_k=H,a_k=a]
-\mathbb E_{\widetilde{\mathcal P}(\cdot\mid B,a)}[\tilde V^{\widetilde\pi^*}(B_{k+1})]\bigr|\Bigr\}.
\end{aligned}
\]
The reward, continuation-value, and transition terms are bounded separately as follows:
\[
\begin{aligned}
&\bigl|\mathbb E[r_k\mid H_k=H,a_k=a]-\tilde r(B,a)\bigr|
\le 2r_{\max}(1-\rho)^\ell,\\
&\mathbb E\!\left[\bigl|V^{\pi^*}(H_{k+1})-\tilde V^{\widetilde\pi^*}(B_{k+1})\bigr|\mid H_k=H,a_k=a\right]
\le\Delta,\\
&\bigl|\mathbb E[\tilde V^{\widetilde\pi^*}(B_{k+1})\mid H_k=H,a_k=a]
-\mathbb E_{\widetilde{\mathcal P}(\cdot\mid B,a)}[\tilde V^{\widetilde\pi^*}(B_{k+1})]\bigr|\\
&\qquad\le 2\|\tilde V^{\widetilde\pi^*}\|_\infty
\bigl\|\mathbb P(B_{k+1}\in\cdot\mid H_k=H,a_k=a)-\widetilde{\mathcal P}(\cdot\mid B,a)\bigr\|_{\mathrm{TV}}\\
&\qquad\le 2\|\tilde V^{\widetilde\pi^*}\|_\infty(1-\rho)^\ell.
\end{aligned}
\]
The reward bound follows from Lemma~\ref{lem:reward-transition-mismatch}. The continuation-value bound follows from the definition of $\Delta$. For the transition term, we use $|\mathbb E_P[f]-\mathbb E_Q[f]|\le 2\|f\|_\infty\|P-Q\|_{\mathrm{TV}}$ and marginalize the next action in Lemma~\ref{lem:reward-transition-mismatch}. Substituting these bounds yields
\[
\bigl|V^{\pi^*}(H)-\tilde V^{\widetilde\pi^*}(B)\bigr|
\le 2r_{\max}(1-\rho)^\ell+\gamma\Delta
+2\gamma\|\tilde V^{\widetilde\pi^*}\|_\infty(1-\rho)^\ell.
\]
Since $\|\tilde V^{\widetilde\pi^*}\|_\infty\le r_{\max}/(1-\gamma)$, taking the supremum over $H$ gives
\[
\Delta\le\gamma\Delta+2r_{\max}(1-\rho)^\ell
+\frac{2\gamma r_{\max}}{1-\gamma}(1-\rho)^\ell
=\gamma\Delta+\frac{2r_{\max}}{1-\gamma}(1-\rho)^\ell.
\]
Rearranging proves the desired bound for every $H$ and $B=\mathcal G(H)$:
\[
\bigl|V^{\pi^*}(H)-\tilde V^{\widetilde\pi^*}(B)\bigr|
\le\Delta\le\frac{2r_{\max}}{(1-\gamma)^2}(1-\rho)^\ell.\qedhere
\]
\end{proof}

{\color{black}
\section{Full-Reset Occupancy and Stationarity}
\label{app:occupancy-stationarity}
\begin{lemma}[Exact occupancy-stationarity identity]
\label{lem:occupancy-stationarity}
Fix a superstate policy $\pi$. Let
\[
d^\pi(B)
:=
(1-\gamma)\sum_{t=0}^{\infty}\gamma^t
\mathbb P(B_t=B\mid B_0=\varnothing,\pi)
\]
be the normalized discounted visitation distribution of the original superstate process. Then $d^\pi$ is the unique stationary distribution of the full-reset process
\[
\mathcal P^\pi(B'\mid B)
=
(1-\gamma)\delta_{\varnothing}(B')
+
\gamma\sum_{a\in\mathcal A}\pi(a\mid B)
\widetilde{\mathcal P}(B'\mid B,a).
\]
Consequently, $\mu^\pi(B,a):=d^\pi(B)\pi(a\mid B)$ is the stationary distribution of the superstate-action chain with full resets. 
\end{lemma}

\begin{proof}
All time-indexed probabilities below refer to the original superstate process without resets. First, $d^\pi$ is normalized because
\[
\begin{aligned}
\sum_{B\in\mathsf B}d^\pi(B)
&=(1-\gamma)\sum_{t=0}^{\infty}\gamma^t\sum_{B\in\mathsf B}\mathbb P(B_t=B\mid B_0=\varnothing,\pi)\\
&=(1-\gamma)\sum_{t=0}^{\infty}\gamma^t=1.
\end{aligned}
\]
For every $t\geq0$, the Markov property of the original SMDP and the law of total probability give
\[
\mathbb P(B_{t+1}=B'\mid B_0=\varnothing,\pi)
=\sum_{B\in\mathsf B}\mathbb P(B_t=B\mid B_0=\varnothing,\pi)\sum_{a\in\mathcal A}\pi(a\mid B)\widetilde{\mathcal P}(B'\mid B,a).
\]
This sums over the current superstate and the action selected there. Substituting the restart kernel separates the reset and non-reset terms. Substituting $d^\pi$ into the latter term and applying the preceding identity then yields
\[
\begin{aligned}
\sum_{B\in\mathsf B}d^\pi(B)\mathcal P^\pi(B'\mid B)
&=(1-\gamma)\delta_{\varnothing}(B')\sum_{B\in\mathsf B}d^\pi(B)+\gamma\sum_{B\in\mathsf B}\sum_{a\in\mathcal A}d^\pi(B)\pi(a\mid B)\widetilde{\mathcal P}(B'\mid B,a)\\
&=
(1-\gamma)\delta_{\varnothing}(B')
+
\gamma(1-\gamma)
\sum_{t=0}^{\infty}\gamma^t
\mathbb P(B_{t+1}=B'\mid B_0=\varnothing,\pi)\\
&=
(1-\gamma)\delta_{\varnothing}(B')
+
(1-\gamma)
\sum_{t=1}^{\infty}\gamma^t
\mathbb P(B_t=B'\mid B_0=\varnothing,\pi)\\
&=(1-\gamma)\sum_{t=0}^{\infty}\gamma^t\mathbb P(B_t=B'\mid B_0=\varnothing,\pi)\\
&=d^\pi(B').
\end{aligned}
\]
Hence $d^\pi$ is stationary for $\mathcal P^\pi$.
Since the next action is sampled from $\pi(\cdot\mid B')$, the same identity gives stationarity of $\mu^\pi(B,a)=d^\pi(B)\pi(a\mid B)$ for the state-action process.

It remains to prove uniqueness. Let $\lambda$ and $\lambda'$ be any two probability distributions on $\mathsf B$.
For this superstate chain, write $\widetilde{\mathcal P}^\pi(B'\mid B):=\sum_{a\in\mathcal A}\pi(a\mid B)\widetilde{\mathcal P}(B'\mid B,a)$.
The reset term is independent of $B$ and cancels because both distributions sum to one.
\[
\bigl\|{\textstyle\sum_{B\in\mathsf B}}(\lambda(B)-\lambda'(B))\,
\mathcal P^\pi(\cdot\mid B)\bigr\|_{\mathrm{TV}}
=\gamma\bigl\|{\textstyle\sum_{B\in\mathsf B}}(\lambda(B)-\lambda'(B))\,
\widetilde{\mathcal P}^\pi(\cdot\mid B)\bigr\|_{\mathrm{TV}}
\leq\gamma\|\lambda-\lambda'\|_{\mathrm{TV}}.
\]
The inequality follows from the nonexpansiveness of a Markov kernel in total variation.
If $d_1$ and $d_2$ were stationary, then
\[
\|d_1-d_2\|_{\mathrm{TV}}
=\bigl\|{\textstyle\sum_{B\in\mathsf B}}(d_1(B)-d_2(B))\,
\mathcal P^\pi(\cdot\mid B)\bigr\|_{\mathrm{TV}}
\le\gamma\|d_1-d_2\|_{\mathrm{TV}}.
\]
Since $\gamma<1$, this implies $d_1=d_2$.
\end{proof}

\subsection{Proof of Lemma~\ref{lem:full-reset-mixing}}
\begin{proof}
For the full-reset superstate-action chain, define the transition kernel
\[
\mathcal P^\pi(B',a'\mid B,a)
:=
\left[(1-\gamma)\delta_{\varnothing}(B')
+\gamma\widetilde{\mathcal P}(B'\mid B,a)\right]\pi(a'\mid B').
\]
Here $\mathcal P^\pi$ acts on $\mathsf B\times\mathcal A$. Its arguments distinguish it from the superstate kernel $\mathcal P^\pi(B'\mid B)$ in Lemma~\ref{lem:occupancy-stationarity}.
By Lemma~\ref{lem:occupancy-stationarity}, $\mu^\pi(B,a)=d^\pi(B)\pi(a\mid B)$ is stationary for $\mathcal P^\pi$.
Let $\lambda$ and $\lambda'$ be any two probability distributions on $\mathsf B\times\mathcal A$.
Define
\[
\nu^\pi(B',a'):=\delta_{\varnothing}(B')\pi(a'\mid\varnothing),
\qquad
\widetilde{\mathcal P}^\pi(B',a'\mid B,a)
:=\widetilde{\mathcal P}(B'\mid B,a)\pi(a'\mid B').
\]
Then $\mathcal P^\pi=(1-\gamma)\nu^\pi+\gamma\widetilde{\mathcal P}^\pi$, where the reset distribution $\nu^\pi$ is independent of the current state-action pair. Therefore,
\[
\bigl\|{\textstyle\sum_{B,a}}(\lambda-\lambda')(B,a)\,
\mathcal P^\pi(\cdot\mid B,a)\bigr\|_{\mathrm{TV}}
=\gamma\bigl\|{\textstyle\sum_{B,a}}(\lambda-\lambda')(B,a)\,
\widetilde{\mathcal P}^\pi(\cdot\mid B,a)\bigr\|_{\mathrm{TV}}
\leq\gamma\|\lambda-\lambda'\|_{\mathrm{TV}},
\]
where the inequality follows from the nonexpansiveness of a Markov kernel in total variation.
This contraction also implies that $\mu^\pi$ is the unique stationary distribution of $\mathcal P^\pi$.
For each $(B,a)\in\mathsf B\times\mathcal A$, the initial distribution concentrated at $(B,a)$ has TV distance at most one from $\mu^\pi$.
Iterating the contraction and using stationarity gives, for every $k\geq0$,
\[
\left\|\mathbb P\bigl((\tilde B_k,\tilde a_k)\in\cdot\mid
(\tilde B_0,\tilde a_0)=(B,a)\bigr)-\mu^\pi\right\|_{\mathrm{TV}}
\leq\gamma^k.
\]
Taking the supremum over $(B,a)$ proves Lemma~\ref{lem:full-reset-mixing}. The bound is uniform over policies and memory lengths.
\end{proof}
}

{\color{black}
\subsection{The actual full-reset POMDP sampling chain}
\label{app:actual-reset}
\begin{lemma}[Actual mixing and SMDP occupancy mismatch]
\label{lem:actual-reset}
Fix $\pi(\cdot\mid B)$ during a critic run. The actual process $X_k=(S_k,B_k,a_k)$ is a finite Markov chain under the joint reset protocol. Write $\xi^\pi$ for its stationary distribution and $\mu_{\mathrm P}^\pi$ for its $(B,a)$ marginal. Then
\[
\sup_x\|\mathbb P(X_k\in\cdot\mid X_0=x)-\xi^\pi\|_{\mathrm{TV}}\leq\gamma^k.
\]
Under Lemma~\ref{lem:reward-transition-mismatch},
\[
\|\mu_{\mathrm P}^\pi-\mu^\pi\|_{\mathrm{TV}}\leq
\min\left\{1,\frac{\gamma}{1-\gamma}(1-\rho)^\ell\right\}.
\]
\end{lemma}
\begin{proof}
\emph{Mixing of the actual sampling chain.}
On a reset, draw $S'\sim\mathcal D$ and set $B'=\varnothing$. Then draw $a'\sim\pi(\cdot\mid\varnothing)$.
Without a reset, generate the next state and observation using $P$ and $\Phi$. Update the window and sample the next action from $\pi(\cdot\mid B')$.
Thus, the transition kernel on $(s,B,a)$ is
\[
\resizebox{\linewidth}{!}{$\displaystyle
\mathcal P_{\mathrm P}^\pi(s',B',a'\mid s,B,a)
=(1-\gamma)\mathcal D(s')\delta_{\varnothing}(B')\pi(a'\mid\varnothing)
+\gamma P(s'\mid s,a)\pi(a'\mid B')\sum_o\Phi(o\mid s')\mathbf1\{\mathcal G(B\|(a,o))=B'\}.
$}
\]
The reset distribution $\mathcal D(s')\delta_{\varnothing}(B')\pi(a'\mid\varnothing)$ does not depend on the previous $(s,B,a)$.
For any two probability distributions $\lambda$ and $\lambda'$ on the augmented state space, their reset contributions satisfy
\[
(1-\gamma)\mathcal D(s')\delta_{\varnothing}(B')\pi(a'\mid\varnothing)
\left(\sum_x\lambda(x)-\sum_x\lambda'(x)\right)=0.
\]
The remaining transition follows the original dynamics with probability $\gamma$. Applying the TV nonexpansiveness argument in Lemma~\ref{lem:occupancy-stationarity} gives
\[
\left\|\sum_x(\lambda(x)-\lambda'(x))\,
\mathcal P_{\mathrm P}^\pi(\cdot\mid x)\right\|_{\mathrm{TV}}
\leq\gamma\|\lambda-\lambda'\|_{\mathrm{TV}}.
\]
A stationary distribution $\xi^\pi$ exists because the augmented state space is finite. The contraction makes it unique.
For each initial state $x$, the initial distribution concentrated at $x$ has TV distance at most one from $\xi^\pi$.
Iterating the contraction and using stationarity gives
\[
\|\mathbb P(X_k\in\cdot\mid X_0=x)-\xi^\pi\|_{\mathrm{TV}}
\leq\gamma^k.
\]

\emph{Comparison of the original processes.}
For the following comparison, both processes run without resets. The subscripts $\mathrm P$ and $\mathrm{SMDP}$ identify probabilities under the actual POMDP and the SMDP, respectively. Write $\mathbb P_{\mathrm P}((B_j,a_j)\in\cdot)$ and $\mathbb P_{\mathrm{SMDP}}((B_j,a_j)\in\cdot)$ for their window-action distributions.
Both start with $B_0=\varnothing$ and $a_0\sim\pi(\cdot\mid\varnothing)$. The POMDP additionally starts with $S_0\sim\mathcal D$.
Recall from Lemma~\ref{lem:reward-transition-mismatch} that the SMDP superstate-action kernel is
\[
\widetilde{\mathcal P}^\pi(B',a'\mid B,a)
=\widetilde{\mathcal P}(B'\mid B,a)\pi(a'\mid B').
\]
Propagating the current POMDP window-action distribution through one SMDP transition gives the following probability at each $(B',a')$:
\[
\sum_{B,a}\mathbb P_{\mathrm P}(B_j=B,a_j=a)\widetilde{\mathcal P}(B'\mid B,a)\pi(a'\mid B').
\]
This propagated distribution need not be the actual next POMDP distribution.
By Lemma~\ref{lem:reward-transition-mismatch}, the next-pair distribution conditional on each actual history and action differs from $\widetilde{\mathcal P}^\pi(\cdot\mid B_j,a_j)$ by at most $(1-\rho)^\ell$ in TV.
Since $B_j=\mathcal G(H_j)$, averaging over those histories and actions gives
\[
\begin{aligned}
&\left\|\mathbb P_{\mathrm P}((B_{j+1},a_{j+1})\in\cdot)
-\sum_{B,a}\mathbb P_{\mathrm P}(B_j=B,a_j=a)\widetilde{\mathcal P}^\pi(\cdot\mid B,a)\right\|_{\mathrm{TV}}\\
&\quad=\left\|\mathbb E_{\mathrm P}\!\left[\mathbb P_{\mathrm P}((B_{j+1},a_{j+1})\in\cdot\mid H_j,a_j)-\widetilde{\mathcal P}^\pi(\cdot\mid B_j,a_j)\right]\right\|_{\mathrm{TV}}\\
&\quad\leq\mathbb E_{\mathrm P}\!\left[\left\|\mathbb P_{\mathrm P}((B_{j+1},a_{j+1})\in\cdot\mid H_j,a_j)-\widetilde{\mathcal P}^\pi(\cdot\mid B_j,a_j)\right\|_{\mathrm{TV}}\right]\\
&\quad\leq(1-\rho)^\ell.
\end{aligned}
\]
The equality uses the law of total probability. The first inequality uses convexity of TV and the second uses Lemma~\ref{lem:reward-transition-mismatch}.
The SMDP evolves exactly according to its Markov kernel:
\[
\mathbb P_{\mathrm{SMDP}}((B_{j+1},a_{j+1})\in\cdot)
=\sum_{B,a}\mathbb P_{\mathrm{SMDP}}(B_j=B,a_j=a)\widetilde{\mathcal P}^\pi(\cdot\mid B,a).
\]
The triangle inequality and TV nonexpansiveness then give
\[
\begin{aligned}
&\left\|\mathbb P_{\mathrm P}((B_{j+1},a_{j+1})\in\cdot)-\mathbb P_{\mathrm{SMDP}}((B_{j+1},a_{j+1})\in\cdot)\right\|_{\mathrm{TV}}\\
&\quad\leq(1-\rho)^\ell+\left\|\mathbb P_{\mathrm P}((B_j,a_j)\in\cdot)-\mathbb P_{\mathrm{SMDP}}((B_j,a_j)\in\cdot)\right\|_{\mathrm{TV}}.
\end{aligned}
\]
The initial window-action distributions coincide. Induction therefore yields
\[
\left\|\mathbb P_{\mathrm P}((B_j,a_j)\in\cdot)-\mathbb P_{\mathrm{SMDP}}((B_j,a_j)\in\cdot)\right\|_{\mathrm{TV}}
\leq j(1-\rho)^\ell.
\]
Only the SMDP window-action process is treated as Markov in this comparison. The POMDP bound is obtained by conditioning on full histories.

\emph{From time marginals to stationary distributions.}
Apply the occupancy-stationarity calculation of Lemma~\ref{lem:occupancy-stationarity} to $\mathcal P_{\mathrm P}^\pi$ and then marginalize out the latent state. Together with the SMDP identity in that lemma, this gives
\[
\begin{aligned}
\mu_{\mathrm P}^\pi&=(1-\gamma)\sum_{j=0}^{\infty}\gamma^j\mathbb P_{\mathrm P}((B_j,a_j)\in\cdot),\\
\mu^\pi&=(1-\gamma)\sum_{j=0}^{\infty}\gamma^j\mathbb P_{\mathrm{SMDP}}((B_j,a_j)\in\cdot).
\end{aligned}
\]
Hence,
\[
\begin{aligned}
\|\mu_{\mathrm P}^\pi-\mu^\pi\|_{\mathrm{TV}}
&\leq(1-\gamma)\sum_{j=0}^{\infty}\gamma^j\left\|\mathbb P_{\mathrm P}((B_j,a_j)\in\cdot)-\mathbb P_{\mathrm{SMDP}}((B_j,a_j)\in\cdot)\right\|_{\mathrm{TV}}\\
&\leq(1-\gamma)\sum_{j=0}^{\infty}\gamma^j j(1-\rho)^\ell\\
&=\frac{\gamma}{1-\gamma}(1-\rho)^\ell.
\end{aligned}
\]
The TV distance between two probability distributions is also at most one. Combining the two bounds proves the stated minimum.
\end{proof}
}

\section{Proof of Critic Convergence}
\subsection{Preliminaries}
\paragraph{{\color{black}Goal for Critic.}}
Fix a policy $\pi$. Let $\widehat Q_k(\cdot,\cdot)=\widehat Q(\cdot,\cdot;\bar W_k)$ denote the critic estimate produced at iteration $k$, let $\tilde Q^\pi$ be the Bellman fixed point of the 
original superstate MDP
under $\pi$, and let $\mu^\pi$ denote the corresponding stationary state-action distribution of the superstate MDP with full resets (which is also the discounted visitation distribution of original superstate MDP). Our goal is to bound the expected critic error under $\mu^\pi$:
\begin{equation}
\mathbb E\big[\|\widehat Q_k-\tilde Q^\pi\|_{\mu^\pi}^2\big]
=\mathbb E\Big[\mathbb E_{(B,a)\sim\mu^\pi}\big[(\widehat Q_k(B,a)-\tilde Q^\pi(B,a))^2\big]\Big].
\label{goal}
\end{equation}
The main difficulty is that $\widehat W_k$ is not updated using samples drawn directly from the induced superstate MDP. Instead, the critic $\widehat Q_k(\cdot,\cdot)$ is trained using trajectories generated by the underlying POMDP. We evaluate the error under $\mu^\pi$, because $\tilde Q^\pi$ is the Bellman fixed point of the superstate MDP and $\mu^\pi$ is the stationary distribution naturally associated with this target.

\paragraph{Data-generating model for $\widehat{Q}_k$.}
The estimate $\widehat{Q}_k$ is obtained from trajectory samples $(H_k,a_k,r_k,H_{k+1},a_{k+1})$, which are mapped into the superstate representation $(B_k,a_k,r_k,B_{k+1},a_{k+1})$ with $B_k:=\mathcal G(H_k)$. These samples are generated by the underlying POMDP dynamics rather than by the superstate MDP kernel. The reward expectation and next-superstate transition induced by the POMDP are given in Lemma~\ref{lem:reward-transition-mismatch}.

\paragraph{Locally linearized critic and projected Bellman fixed point.}
Following \citet{cai2019neural}, we introduce the locally linearized critic $\widehat{Q}_0$ around the initialization $\widehat{W}_0$:
\[
\widehat{Q}_0(B,a;W):=\widehat{Q}(B,a;\widehat{W}_0)+\big\langle \nabla_W \widehat{Q}(B,a;\widehat{W}_0),\, W-\widehat{W}_0 \big\rangle.
\]

Let $\mathcal{B}:=\left\{W:\|W^{(h)}-\widehat{W}_0^{(h)}\|_F\leq R,\forall h\in[d]\right\}$ and $\mathcal{F}:=\{\widehat{Q}_0(\cdot,\cdot;W):W\in\mathcal{B}\}$, then $\|W-\widehat{W}_0\|_2^2=\sum_{h=1}^d\|W^{(h)}-\widehat{W}_0^{(h)}\|_F^2\leq dR^2$. Let $\Pi_{\mathcal F}$ denote the projection operator onto $\mathcal F$, defined by $\Pi_{\mathcal F} f := \arg\min_{g\in\mathcal F} \|f-g\|_{\mu^\pi}^2$. 
Lemma~\ref{lem:occupancy-stationarity} gives $\mu^\pi(B,a)=(1-\gamma)\delta_{\varnothing}(B)\pi(a\mid\varnothing)+\gamma(\mu^\pi\widetilde{\mathcal P}^\pi)(B,a)$. Hence $\mu^\pi\widetilde{\mathcal P}^\pi\leq\mu^\pi/\gamma$. Since $\mathcal F$ is closed and convex, projection nonexpansiveness and Jensen's inequality yield
\[
\begin{aligned}
\|\Pi_{\mathcal F}\tilde T^\pi f-\Pi_{\mathcal F}\tilde T^\pi g\|_{\mu^\pi}^2
&\leq\|\tilde T^\pi f-\tilde T^\pi g\|_{\mu^\pi}^2\\
&\leq\gamma^2\|f-g\|_{\mu^\pi\widetilde{\mathcal P}^\pi}^2\\
&\leq\gamma\|f-g\|_{\mu^\pi}^2.
\end{aligned}
\]
Taking square roots shows that $\Pi_{\mathcal F}\tilde T^\pi$ is a $\sqrt{\gamma}$-contraction in $L^2(\mu^\pi)$.
Therefore, $Q_{\mathcal F}^{*,\pi}$ is the unique projected Bellman fixed point in $\mathcal{F}$ which satisfies $Q_{\mathcal F}^{*,\pi}=\Pi_{\mathcal F}\tilde{T}^\pi Q_{\mathcal F}^{*,\pi}$, and there exists $W^*\in\mathcal B$ such that $\widehat{Q}_0(\cdot,\cdot;W^*)=Q_{\mathcal F}^{*,\pi}$.

\begin{fact}
\label{fact:local-linearization}
According to \citet{cai2019neural}'s Lemmas F.1 and F.2, let $m=\Omega{\left(\frac{p^{3/2}\log^{3/2}(m^{1/2}/R)}{Rd^{3/2}}\right)},R=O{\left(m^{1/2}d^{-6}\log^{-3}m\right)}$, with probability at least $1-e^{-\Omega(\log^2m)}$ over the random initialization, the following hold for every admissible superstate-action pair $(B,a)$ satisfying $\|\psi(B,a)\|_2=1$ and every $W\in\mathcal{B}:$

\[
\begin{aligned}
\|\nabla_W \widehat{Q}(B,a;W)-\nabla_W \widehat{Q}_0(B,a;W)\|_2 &= O\left(R^{1/3}m^{-1/6}d^{5/2}\log^{1/2}m\right) \\
\|\nabla_W \widehat{Q}(B,a;W)\|_2 &= O(d) \\
|\widehat{Q}(B,a;\widehat W_0)| &= O(\log m) \\
|\widehat{Q}(B,a;W)-\widehat{Q}_0(B,a;W)| &= O\left(R^{4/3}m^{-1/6}d^3\log^{1/2}m\right)
\end{aligned}
\]
\end{fact}

Fact~\ref{fact:local-linearization} depends only on the neural network architecture and the input feature $x=\psi(B,a)$, through the condition $\|x\|_2=1$, and is independent of the distribution from which $(B,a)$ is sampled to input and calculate $W$.

\begin{lemma}[Lipschitz and bounded critic]
\label{lem:lipschitz-critic}
Under the same condition of Fact~\ref{fact:local-linearization}, with probability at least $1-e^{-\Omega(\log^2m)}$ over the random initialization, there exist $L_Q = O(d)$ and $Q_{\max} = O(\log m + d^{3/2}R)$ such that for all $W, W' \in \mathcal{B}$ and $(B, a)$ satisfying $\|\psi(B,a)\|_2=1$:
\begin{equation*}
\begin{aligned}
\|\nabla_W \widehat{Q}(B, a; W)\|_2 &\le L_Q, \\
|\widehat{Q}(B, a; W) - \widehat{Q}(B, a; W')| &\le L_Q \|W - W'\|_2, \\
|\widehat{Q}_0(B, a; W) - \widehat{Q}_0(B, a; W')| &\le L_Q \|W - W'\|_2, \\
\max\{|\widehat{Q}(B, a; W)|, |\widehat{Q}_0(B, a; W)|\} &\le Q_{\max}.
\end{aligned}
\end{equation*}
\end{lemma}

\begin{proof}
By Fact~\ref{fact:local-linearization}, $\|\nabla_W \widehat{Q}(B, a; W)\|_2 = O(d)$, which allows us to set $L_Q = \sup_{W \in \mathcal{B}, (B, a)} \|\nabla_W \widehat{Q}(B, a; W)\|_2$. 
Since $\mathcal{B}$ is convex, we have
$|\widehat{Q}(B, a; W) - \widehat{Q}(B, a; W')| \le L_Q \|W - W'\|_2.$
For the linearized critic $\widehat{Q}_0$, the difference can be expressed and bounded as follows:
\begin{align*}
|\widehat{Q}_0(B, a; W) - \widehat{Q}_0(B, a; W')| &= |\langle \nabla_W \widehat{Q}(B, a; \widehat{W}_0), W - W' \rangle| \\
&\le \|\nabla_W \widehat{Q}(B, a; \widehat{W}_0)\|_2 \|W - W'\|_2 \\
&\le L_Q \|W - W'\|_2,
\end{align*}

For boundedness, Fact~\ref{fact:local-linearization} implies
$|\widehat Q(B,a;\widehat W_0)|=O(\log m)$. Given $W\in\mathcal B$, we have
$\|W-\widehat W_0\|_2\le \sqrt d R$. Hence, by the Lipschitz property,
\[
\begin{aligned}
|\widehat Q(B,a;W)|
&\le |\widehat Q(B,a;\widehat W_0)|+L_Q\|W-\widehat W_0\|_2 \\
&\le O(\log m)+O(d)\cdot \sqrt d R \\
&=O(\log m+d^{3/2}R).
\end{aligned}
\]
Similarly, by the linearized definition
$
\widehat Q_0(B,a;W)
=\widehat Q(B,a;\widehat W_0)
+\bigl\langle\nabla_W\widehat Q(B,a;\widehat W_0),\,W-\widehat W_0\bigr\rangle,
$
we have
\[
\begin{aligned}
|\widehat Q_0(B,a;W)|
&\le |\widehat Q(B,a;\widehat W_0)|
+\|\nabla_W\widehat Q(B,a;\widehat W_0)\|_2\|W-\widehat W_0\|_2 \\
&\le O(\log m)+O(d)\cdot \sqrt d R \\
&=O(\log m+d^{3/2}R).
\end{aligned}
\]
which completes the proof by taking $Q_{\max} = O(\log m + d^{3/2}R)$.
\end{proof}

\subsection{Proof of Theorem \ref{thm:critic-error}}
\label{app:critic}
We decompose the error in equation \ref{goal} by:
\begin{equation}
\label{eq:critic-error-decomp}
\begin{aligned}
\mathbb E\big[\|\widehat Q(\cdot,\cdot;\bar W_K)-\tilde Q^\pi\|_{\mu^\pi}^2\big]
&\le 3\mathbb E\underbrace{\big[\|\widehat Q(\cdot,\cdot;\bar W_K)-\widehat Q_0(\cdot,\cdot;\bar W_K)\|_{\mu^\pi}^2\big]}_{\text{NTK linearization error}}\\
&\quad+3\mathbb E\underbrace{\big[\|\widehat Q_0(\cdot,\cdot;\bar W_K)-Q_{\mathcal F}^{*,\pi}\|_{\mu^\pi}^2\big]}_{\text{projected optimization error}}\\
&\quad+3\mathbb E\big [\underbrace{\|Q_{\mathcal F}^{*,\pi}-\tilde Q^\pi\|_{\mu^\pi}^2}_{\text{approximation error}}\big].
\end{aligned}
\end{equation}

\subsubsection{Bound NTK linearization error}
Since each iterate $\widehat W_k\in\mathcal B$ and $\mathcal B$ is convex, we have $\bar W_K\in\mathcal B$. Using Fact~\ref{fact:local-linearization} with $W=\bar W_K$, we obtain that, with probability at least $1-e^{-\Omega(\log^2 m)}$,
\[
\mathbb E\big[\|\widehat Q(\cdot,\cdot;\bar W_K)-\widehat Q_0(\cdot,\cdot;\bar W_K)\|_{\mu^\pi}^2\big]=O\!\left(R^{8/3}m^{-1/3}d^6\log m\right).
\]
\subsubsection{Bound projected optimization error}
\label{app:critic-scheme-b}

\paragraph{\textbf{Step 1. Lyapunov Analysis.}}

We use Lyapunov analysis to control $\|\widehat{Q}_0(\cdot,\cdot;\widehat{W}_k)-\widehat{Q}_0(\cdot,\cdot;W^*)\|_{\mu^\pi}^2$.
Define $e_k:=\widehat{W}_k-W^*$.
The projection ball gives $\|e_k\|_2\leq2\sqrt d\,R$.
Due to $\widehat{W}_{k+1}=\mathrm{Proj}_{\mathcal{B}}(\widehat{W}_k-\eta g_k(\widehat{W}_k))$ and the non-expansion property of projection, we can get
\begin{equation}
\|e_{k+1}\|_2^2
\leq
\|e_k\|_2^2-2\eta\langle g_k(\widehat{W}_k),e_k\rangle+\eta^2\|g_k(\widehat{W}_k)\|_2^2.
\label{eq:lyapunov-recursion-raw-pre-reset}
\end{equation}
where the true POMDP stochastic semigradient $g_k(\widehat{W}_k)$ is defined by

{\color{black}
\[
g_k(W):=Z_{k+1}\delta_{k}(W)\nabla_W\widehat{Q}(B_k,a_k;W),
\]
}
\[
\delta_{k}(W):=\widehat{Q}(B_k,a_k;W)-[r_k+\gamma \widehat{Q}(B_{k+1},a_{k+1};W)].
\]

{\color{black}
Here $Z_{k+1}$ is the Bernoulli variable in Algorithm~\ref{alg:neural_superstate_td}.
}
{\color{black}
The displayed TD residual is evaluated only when $Z_{k+1}=1$ and the successor follows the original POMDP dynamics.
}
{\color{black}
On $Z_{k+1}=0$, set both TD residuals $\delta_k(W)$ and $\delta_{0,k}(W)$ to zero for notation.
}
{\color{black}
No critic update is performed on that branch.
}

According to Lemma~\ref{lem:lipschitz-critic} and Assumption~\ref{assump:bounded-reward},
\[
\|g_k(\widehat{W}_k)\|_2^2
\leq
\left(r_{\max}+(1+\gamma)Q_{\max}\right)^2L_Q^2
=:\delta_{\max}^2L_Q^2,\qquad
\delta_{\max}=O(\log m+d^{3/2}R).
\]
For the analysis, we further introduce two semigradients based on the
locally linearized critic $\widehat{Q}_0$.
The first is the linearized POMDP semigradient

{\color{black}
\[
g_{0,k}(W)
=
Z_{k+1}\delta_{0,k}(W)\nabla_W \widehat{Q}_0(B_k,a_k),
\]
}
\[
\delta_{0,k}(W)
=
\widehat{Q}_0(B_k,a_k;W)-[r_k+\gamma \widehat{Q}_0(B_{k+1},a_{k+1};W)].
\]
The second is the linearized semigradient under the induced superstate MDP:
\[
g_{0}^{\mathrm{SMDP}}(B,a;W)
:=
\delta_{0}^{\mathrm{SMDP}}(B,a;W)\nabla_W \widehat{Q}_0(B,a),
\]
\[
\delta_{0}^{\mathrm{SMDP}}(B,a;W)
:=
\widehat{Q}_0(B,a;W)
-
\left[
\tilde{r}(B,a)
+
\gamma
\sum_{B^{\prime},a^{\prime}}
\widetilde{\mathcal{P}}^\pi(B^{\prime},a^{\prime}\mid B,a)\widehat{Q}_0(B^{\prime},a^{\prime};W)
\right].
\]

Define the corresponding population semigradient
\[
\bar g_0^{\mathrm{SMDP}}(W)
:=
\mathbb E_{(B,a)\sim\mu^\pi}\!\left[g_{0}^{\mathrm{SMDP}}(B,a;W)\right].
\]

{\color{black}
Let $\mathcal I_k$ contain the initialization and the augmented chain $X_j=(S_j,B_j,a_j)$ through time $k$.
}
{\color{black}
It also contains the past rewards and the critic iterates.
}
{\color{black}
This filtration is used only in the analysis.
}
{\color{black}
Define the actual stationary population direction by
}
{\color{black}
\[
\bar g_0^{\mathrm{POMDP}}(W):=
\mathbb E_{X_k\sim\xi^\pi}\!\left[\mathbb E\!\left[\delta_{0,k}(W)\nabla_W\widehat Q_0(B_k,a_k)\mid X_k,\ Z_{k+1}=1\right]\right].
\]
}
{\color{black}
Here $\xi^\pi$ is the stationary law from Lemma~\ref{lem:actual-reset}.
}
{\color{black}
This expectation uses a stationary current state and one original POMDP transition.
}
{\color{black}
Set $G_{\max}:=\delta_{\max}L_Q$.
}
{\color{black}
Independence of the reset coin gives
}
{\color{black}
\[
\mathbb E[\|g_k(\widehat W_k)\|_2^2\mid\mathcal I_k]\leq\gamma G_{\max}^2.
\]
}
{\color{black}
We compare the two population directions before controlling temporal dependence.
}

{\color{black}
For the full-reset sampling scheme, we decompose
}

{\color{black}
\begin{equation}
\label{eq:gradient-decomposition-pre-reset}
\begin{aligned}
g_k(\widehat W_k)-\gamma\bar g_0^{\mathrm{SMDP}}(\widehat W_k)
&=\underbrace{\bigl(g_k(\widehat W_k)-g_{0,k}(\widehat W_k)\bigr)}_{\nu_k\text{ (linearization error)}}\quad+\underbrace{\gamma\bigl(\bar g_0^{\mathrm{POMDP}}(\widehat W_k)-\bar g_0^{\mathrm{SMDP}}(\widehat W_k)\bigr)}_{\omega_k\text{ (population mismatch)}}\\
&\quad+\underbrace{\bigl(g_{0,k}(\widehat W_k)-\gamma\bar g_0^{\mathrm{POMDP}}(\widehat W_k)\bigr)}_{\zeta_k\text{ (Markovian sampling error)}}.
\end{aligned}
\end{equation}
}
Substituting \eqref{eq:gradient-decomposition-pre-reset} into
\eqref{eq:lyapunov-recursion-raw-pre-reset} gives

{\color{black}
\begin{equation}
\label{eq:lyapunov-recursion-decomp-pre-reset}
\begin{aligned}
\|e_{k+1}\|_2^2
&\le
\|e_k\|_2^2
-2\eta\gamma\langle \bar g_0^{\mathrm{SMDP}}(\widehat W_k),e_k\rangle
-2\eta\langle \nu_k,e_k\rangle
-2\eta\langle \omega_k,e_k\rangle
\\
&\quad
-2\eta\langle \zeta_k,e_k\rangle
+\eta^2\|g_k(\widehat W_k)\|_2^2.
\end{aligned}
\end{equation}
}

\paragraph{\textbf{Step 2. Main Descent Term $\langle \bar g_0^{\mathrm{SMDP}}(\widehat W_k),e_k\rangle$}.}

\begin{lemma}[Projected stationary point and one-point monotonicity]
\label{fact:one-point-monotonicity}
Let $W^*\in\mathcal B$ satisfy $\widehat{Q}_0(\cdot,\cdot;W^*)=Q_{\mathcal F}^{*,\pi}$.
Then, for all $W\in\mathcal B$,
\[
\langle \bar g_0^{\mathrm{SMDP}}(W^*),\,W-W^*\rangle\ge 0,
\]
and

{\color{black}
\[
\left\langle \bar g_0^{\mathrm{SMDP}}(W)-\bar g_0^{\mathrm{SMDP}}(W^*),\,W-W^*\right\rangle
\ge
(1-\sqrt\gamma)\|\widehat{Q}_0(\cdot,\cdot;W)-\widehat{Q}_0(\cdot,\cdot;W^*)\|_{\mu^\pi}^2.
\]
}
\end{lemma}

\begin{proof}
We first prove the stationary-point inequality.
For any $W\in\mathcal B$, starting from the definition of $\bar g_0^{\mathrm{SMDP}}(W^*)$, we have
\[
\begin{aligned}
\langle \bar g_0^{\mathrm{SMDP}}(W^*),\,W-W^*\rangle
&=
\left\langle \mathbb E_{(B,a)\sim\mu^\pi}\!\left[\delta_0^{\mathrm{SMDP}}(B,a;W^*)\nabla_W\widehat Q_0(B,a)\right],\,W-W^*\right\rangle \\
&=
\mathbb E_{(B,a)\sim\mu^\pi}\!\left[\delta_0^{\mathrm{SMDP}}(B,a;W^*)\langle \nabla_W\widehat Q_0(B,a),\,W-W^*\rangle\right] \\
&=
\mathbb E_{(B,a)\sim\mu^\pi}\!\left[\delta_0^{\mathrm{SMDP}}(B,a;W^*)\bigl(\widehat Q_0(B,a;W)-\widehat Q_0(B,a;W^*)\bigr)\right]\\
&\ge 0.
\end{aligned}
\]
where the last inequality holds since $Q_{\mathcal F}^{*,\pi}=\widehat Q_0(\cdot,\cdot;W^*)$ and $Q_{\mathcal F}^{*,\pi}=\Pi_{\mathcal F}\tilde T^\pi Q_{\mathcal F}^{*,\pi}$, the first-order optimality condition of projection onto the closed convex set $\mathcal F$ yields
\[
\left\langle Q_{\mathcal F}^{*,\pi}-\tilde T^\pi Q_{\mathcal F}^{*,\pi},\,\widehat Q_0(\cdot,\cdot;W)-Q_{\mathcal F}^{*,\pi}\right\rangle_{\mu^\pi}\ge 0.
\]
Using
$
Q_{\mathcal F}^{*,\pi}(B,a)-(\tilde T^\pi Q_{\mathcal F}^{*,\pi})(B,a)=\delta_0^{\mathrm{SMDP}}(B,a;W^*),
$ then we are done.

We next prove the one-point monotonicity inequality.
We have
\[
\begin{aligned}
&\left\langle \bar g_0^{\mathrm{SMDP}}(W)-\bar g_0^{\mathrm{SMDP}}(W^*),\,W-W^*\right\rangle \\
&=
\left\langle
\mathbb E_{(B,a)\sim\mu^\pi}\!\left[\bigl(\delta_0^{\mathrm{SMDP}}(B,a;W)-\delta_0^{\mathrm{SMDP}}(B,a;W^*)\bigr)\nabla_W\widehat Q_0(B,a)\right],
\,W-W^*
\right\rangle \\
&=
\mathbb E_{(B,a)\sim\mu^\pi}\!\left[
\bigl(\delta_0^{\mathrm{SMDP}}(B,a;W)-\delta_0^{\mathrm{SMDP}}(B,a;W^*)\bigr)
\langle \nabla_W\widehat Q_0(B,a),\,W-W^*\rangle
\right] \\
&=
\mathbb E_{(B,a)\sim\mu^\pi}\!\Bigl[
\bigl(\delta_0^{\mathrm{SMDP}}(B,a;W)-\delta_0^{\mathrm{SMDP}}(B,a;W^*)\bigr)
\bigl(\widehat Q_0(B,a;W)-\widehat Q_0(B,a;W^*)\bigr)
\Bigr].
\end{aligned}
\]

Next, by the definition of $\delta_0^{\mathrm{SMDP}}$,
\[
\begin{aligned}
&\delta_0^{\mathrm{SMDP}}(B,a;W)-\delta_0^{\mathrm{SMDP}}(B,a;W^*) \\
&=
\widehat Q_0(B,a;W)-\widehat Q_0(B,a;W^*)
-\gamma\sum_{B',a'}\widetilde{\mathcal P}^{\pi}(B',a'|B,a)\bigl(\widehat Q_0(B',a';W)-\widehat Q_0(B',a';W^*)\bigr).
\end{aligned}
\]
Substituting this identity into the previous display yields
\[
\begin{aligned}
&\left\langle \bar g_0^{\mathrm{SMDP}}(W)-\bar g_0^{\mathrm{SMDP}}(W^*),\,W-W^*\right\rangle \\
& =
\mathbb E_{\mu^\pi}\!\left[\bigl(\widehat Q_0(B,a;W)-\widehat Q_0(B,a;W^*)\bigr)^2\right] \\
&\quad-\gamma\,\mathbb E_{\mu^\pi}\!\left[
\bigl(\widehat Q_0(B,a;W)-\widehat Q_0(B,a;W^*)\bigr)\,
\mathbb E\!\left[\widehat Q_0(B',a';W)-\widehat Q_0(B',a';W^*)\,\middle|\,B,a\right]
\right]\\
&=\mathbb E_{\mu^\pi}\!\left[\bigl(\widehat Q_0(B,a;W)-\widehat Q_0(B,a;W^*)\bigr)^2\right] \\
&
\quad -\gamma \mathbb E_{\mu^\pi}\!\left[
\bigl(\widehat Q_0(B,a;W)-\widehat Q_0(B,a;W^*)\bigr)
\bigl(\widehat Q_0(B',a';W)-\widehat Q_0(B',a';W^*)\bigr)
\right]
\end{aligned}
\]
where the last step comes from the tower property.

{\color{black}
The successor follows the original SMDP kernel.
}
{\color{black}
Its marginal is $\mu^\pi\widetilde{\mathcal P}^\pi$.
}
{\color{black}
By Lemma~\ref{lem:occupancy-stationarity},
}
{\color{black}
\[
\mu^\pi(B',a')=(1-\gamma)\delta_{\varnothing}(B')\pi(a'\mid\varnothing)+\gamma(\mu^\pi\widetilde{\mathcal P}^\pi)(B',a'),
\qquad \mu^\pi\widetilde{\mathcal P}^\pi\leq\mu^\pi/\gamma.
\]
}

Therefore, by Cauchy--Schwarz,

{\color{black}
\[
\begin{aligned}
&\mathbb E_{\mu^\pi}\!\left[
\bigl(\widehat Q_0(B,a;W)-\widehat Q_0(B,a;W^*)\bigr)
\bigl(\widehat Q_0(B',a';W)-\widehat Q_0(B',a';W^*)\bigr)
\right]\\
&\leq
\|\widehat Q_0(\cdot,\cdot;W)-\widehat Q_0(\cdot,\cdot;W^*)\|_{\mu^\pi}
\|\widehat Q_0(\cdot,\cdot;W)-\widehat Q_0(\cdot,\cdot;W^*)\|_{\mu^\pi\widetilde{\mathcal P}^\pi}\\
&\leq\gamma^{-1/2}
\|\widehat Q_0(\cdot,\cdot;W)-\widehat Q_0(\cdot,\cdot;W^*)\|_{\mu^\pi}^2.
\end{aligned}
\]
}

Combining the above displays, we obtain

{\color{black}
\[
\left\langle \bar g_0^{\mathrm{SMDP}}(W)-\bar g_0^{\mathrm{SMDP}}(W^*),\,W-W^*\right\rangle
\ge
(1-\sqrt\gamma)\|\widehat Q_0(\cdot,\cdot;W)-\widehat Q_0(\cdot,\cdot;W^*)\|_{\mu^\pi}^2.
\]
}

This completes the proof.
\end{proof}

Applying Lemma~\ref{fact:one-point-monotonicity} at $W=\widehat W_k$, we obtain

{\color{black}
\[
-2\eta\gamma\langle \bar g_0^{\mathrm{SMDP}}(\widehat W_k),e_k\rangle
\le
-2\eta\gamma(1-\sqrt\gamma)
\|\widehat{Q}_0(\cdot,\cdot;\widehat W_k)-\widehat{Q}_0(\cdot,\cdot;W^*)\|_{\mu^\pi}^2.
\]
}

This is crucial because it yields the descent term in the Lyapunov recursion.

\paragraph{\textbf{Step 3. Bound on $\langle\nu_k,e_k\rangle$.}}

Recall that $\nu_k=g_k(\widehat W_k)-g_{0,k}(\widehat W_k)$.

{\color{black}
On a non-reset branch, by definition,
}
\[
\begin{aligned}
\|\nu_k\|_2
&=
\left\|
\delta_k(\widehat W_k)\nabla_W\widehat Q(B_k,a_k;\widehat W_k)
-
\delta_{0,k}(\widehat W_k)\nabla_W\widehat Q_0(B_k,a_k)
\right\|_2 \\
&\le
\left|\delta_k(\widehat W_k)\right|
\left\|\nabla_W\widehat Q(B_k,a_k;\widehat W_k)-\nabla_W\widehat Q_0(B_k,a_k)\right\|_2
\\&\qquad+
\left|\delta_k(\widehat W_k)-\delta_{0,k}(\widehat W_k)\right|
\left\|\nabla_W\widehat Q_0(B_k,a_k)\right\|_2 .
\end{aligned}
\]
Moreover,
\[
\left|\delta_k(\widehat W_k)\right|
\le
r_{\max}+(1+\gamma)Q_{\max}
=:\delta_{\max},
\]
and

\[
\begin{aligned}
\left|\delta_k(\widehat W_k)-\delta_{0,k}(\widehat W_k)\right|
&\le
\left|\widehat Q(B_k,a_k;\widehat W_k)-\widehat Q_0(B_k,a_k;\widehat W_k)\right| \\
&\quad+
\gamma\left|\widehat Q(B_{k+1},a_{k+1};\widehat W_k)-\widehat Q_0(B_{k+1},a_{k+1};\widehat W_k)\right|.
\end{aligned}
\]
Thus, by Fact~\ref{fact:local-linearization}, with probability at least $1-e^{-\Omega(\log^2 m)}$,
\[
\|\nu_k\|_2^2
=
O\!\left(
\delta_{\max}^2R^{2/3}m^{-1/3}d^5\log m
+
L_Q^2R^{8/3}m^{-1/3}d^6\log m
\right).
\]

{\color{black}
On a reset branch $\nu_k=0$.
}
{\color{black}
Applying Young's inequality on the non-reset branch and using the independence of $Z_{k+1}$ gives, for any $\alpha>0$,
}

{\color{black}
\[
\begin{aligned}
-2\eta\mathbb E\langle\nu_k,e_k\rangle
&\leq\eta\gamma\alpha\mathbb E\|e_k\|_2^2\\
&\quad+\frac{\eta\gamma}{\alpha}O\!\left(
\delta_{\max}^2R^{2/3}m^{-1/3}d^5\log m
+R^{8/3}m^{-1/3}d^8\log m
\right).
\end{aligned}
\]
}
{\color{black}
The first term is retained in the Lyapunov recursion and bounded after summation.
}

\paragraph{\color{black}\textbf{Step 4. Bound on $\langle\omega_k,e_k\rangle$.}}

{\color{black}
The term $\omega_k$ captures the stationary population mismatch between the POMDP and SMDP directions.
}

{\color{black}
For fixed $W\in\mathcal B$, insert the SMDP direction averaged under the actual marginal $\mu_{\mathrm P}^\pi$:
}

{\color{black}
\[
\begin{aligned}
\bar g_0^{\mathrm{POMDP}}(W)-\bar g_0^{\mathrm{SMDP}}(W)
&=\underbrace{\Bigl(\bar g_0^{\mathrm{POMDP}}(W)-\mathbb E_{\mu_{\mathrm P}^\pi}[g_0^{\mathrm{SMDP}}(B,a;W)]\Bigr)}_{\text{one-step reward and transition mismatch}}\\
&\quad+\underbrace{\Bigl(\mathbb E_{\mu_{\mathrm P}^\pi}[g_0^{\mathrm{SMDP}}(B,a;W)]-\bar g_0^{\mathrm{SMDP}}(W)\Bigr)}_{\text{occupancy mismatch}}.
\end{aligned}
\]
}

{\color{black}
By Lemma~\ref{lem:reward-transition-mismatch}, averaging over the full histories of stationary POMDP episodes gives
}
{\color{black}
\[
\left\|\bar g_0^{\mathrm{POMDP}}(W)-\mathbb E_{\mu_{\mathrm P}^\pi}[g_0^{\mathrm{SMDP}}(B,a;W)]\right\|_2
\leq2L_Q(r_{\max}+\gamma Q_{\max})(1-\rho)^\ell.
\]
}
{\color{black}
Since $\|g_0^{\mathrm{SMDP}}(B,a;W)\|_2\leq G_{\max}$, Lemma~\ref{lem:actual-reset} bounds the second term.
}
Hence,

{\color{black}
\[
\begin{aligned}
\|\bar g_0^{\mathrm{POMDP}}(W)-\bar g_0^{\mathrm{SMDP}}(W)\|_2
&\leq2L_Q(r_{\max}+\gamma Q_{\max})(1-\rho)^\ell
+2G_{\max}\|\mu_{\mathrm P}^\pi-\mu^\pi\|_{\mathrm{TV}}\\
&\leq\frac{2G_{\max}}{1-\gamma}(1-\rho)^\ell.
\end{aligned}
\]
}

{\color{black}
This bound is uniform over $W\in\mathcal B$.
}
{\color{black}
Substitute $W=\widehat W_k$ and apply Young's inequality separately to the one-step and occupancy differences above.
For any $\alpha>0$, this gives
}

{\color{black}
\[
-2\eta\mathbb E\langle\omega_k,e_k\rangle
\leq2\eta\gamma\alpha\mathbb E\|e_k\|_2^2
+\frac{\eta\gamma}{\alpha}O\!\left(
\frac{L_Q^2\delta_{\max}^2}{(1-\gamma)^2}(1-\rho)^{2\ell}
\right).
\]
}
{\color{black}
The belief comparison above is for a fixed parameter.
}
{\color{black}
It does not condition on the adaptive critic or past rewards.
}

\paragraph{\color{black}\textbf{Step 5. Bound on $\langle\zeta_k,e_k\rangle$.}}

{\color{black}
To control the Markovian bias term, fix an integer $\tau\geq1$.
}
{\color{black}
For $k\geq\tau$, define the older filtration by
\[
\mathcal I_{k-\tau}
:=
\sigma\!\left(
X_0,\ldots,X_{k-\tau},
R_0,\ldots,R_{k-\tau-1},
\widehat W_0,\ldots,\widehat W_{k-\tau}
\right).
\]
}
{\color{black}
We use the actual augmented chain $X_k$ from Step 1.
}
{\color{black}
Lemma~\ref{lem:actual-reset} gives its mixing bound.
}
{\color{black}
For $k\geq\tau$, freeze the critic at $\widehat W_{k-\tau}$.
}
Also recall that
\[
\delta_{\max}:=r_{\max}+(1+\gamma)Q_{\max},\qquad G_{\max}:=\delta_{\max}L_Q,
\]
so that $\|g_k(\widehat W_k)\|_2\le G_{\max}$.

{\color{black}
For the sampled linearized direction, we have
\[
\begin{aligned}
\|g_{0,k}(W)\|_2
&\leq Z_{k+1}\Bigl(|r_k|+|\widehat Q_0(B_k,a_k;W)|
+\gamma|\widehat Q_0(B_{k+1},a_{k+1};W)|\Bigr)
\|\nabla_W\widehat Q_0(B_k,a_k)\|_2\\
&\leq Z_{k+1}\bigl(r_{\max}+(1+\gamma)Q_{\max}\bigr)L_Q
=Z_{k+1}G_{\max}.
\end{aligned}
\]
Taking the stationary expectation conditional on a non-reset transition gives
\[
\|\bar g_0^{\mathrm{POMDP}}(W)\|_2\leq\delta_{\max}L_Q=G_{\max}.
\]
}

{\color{black}
By the definition of $\zeta_k$, we have
}

{\color{black}
\begin{equation}
\label{eq:zeta-lagged-decomp-pre-reset}
\begin{aligned}
\zeta_k
&=\underbrace{\Bigl(g_{0,k}(\widehat W_k)-g_{0,k}(\widehat W_{k-\tau})\Bigr)}_{\text{iterate drift}}\\
&\quad+\underbrace{\Bigl(g_{0,k}(\widehat W_{k-\tau})-\gamma\bar g_0^{\mathrm{POMDP}}(\widehat W_{k-\tau})\Bigr)}_{\text{lagged bias at frozen iterate}}\\
&\quad+\underbrace{\Bigl(\gamma\bar g_0^{\mathrm{POMDP}}(\widehat W_{k-\tau})-\gamma\bar g_0^{\mathrm{POMDP}}(\widehat W_k)\Bigr)}_{\text{population drift}}.
\end{aligned}
\end{equation}
}

\subparagraph{Step 5.1. Lipschitz Continuity of  {\color{black}$g_{0,k}$ and $\bar g_0^{\mathrm{POMDP}}$}.}

For any $W, W' \in \mathcal{B}$, the Temporal Difference (TD) error satisfies

{\color{black}
\[
\begin{aligned}
|\delta_{0,k}(W)-\delta_{0,k}(W')|
&\leq |\widehat Q_0(B_k,a_k;W)-\widehat Q_0(B_k,a_k;W')|\\
&\quad+\gamma|\widehat Q_0(B_{k+1},a_{k+1};W)-\widehat Q_0(B_{k+1},a_{k+1};W')|\\
&\leq(1+\gamma)L_Q\|W-W'\|_2.
\end{aligned}
\]
The bound holds on a non-reset transition. Both residuals are zero on a reset.
}
Hence,

{\color{black}
\[
\begin{aligned}
\|g_{0,k}(W)-g_{0,k}(W')\|_2
&=Z_{k+1}|\delta_{0,k}(W)-\delta_{0,k}(W')|\,
\|\nabla_W\widehat Q_0(B_k,a_k)\|_2\\
&\leq Z_{k+1}(1+\gamma)L_Q^2\|W-W'\|_2.
\end{aligned}
\]
}
Setting $L_0 := (1+\gamma)L_Q^2$,
{\color{black}
the bounds above and conditional Jensen's inequality give
}
{\color{black}
\[
\begin{aligned}
\|g_{0,k}(W)-g_{0,k}(W')\|_2&\leq Z_{k+1}L_0\|W-W'\|_2,\\
\|\bar g_0^{\mathrm{POMDP}}(W)-\bar g_0^{\mathrm{POMDP}}(W')\|_2&\leq L_0\|W-W'\|_2.
\end{aligned}
\]
}
{\color{black}
Projection nonexpansiveness also gives $\|\widehat W_k-\widehat W_{k-\tau}\|_2\leq\eta\tau G_{\max}$.
}

By the triangle inequality and the update   {\color{black}
bound,
} we have

{\color{black}
\[
\begin{aligned}
&\mathbb E\left\|
\Bigl(g_{0,k}(\widehat W_k)-g_{0,k}(\widehat W_{k-\tau})\Bigr)
+\Bigl(\gamma\bar g_0^{\mathrm{POMDP}}(\widehat W_{k-\tau})-\gamma\bar g_0^{\mathrm{POMDP}}(\widehat W_k)\Bigr)
\right\|_2\\
&\leq\mathbb E[(Z_{k+1}+\gamma)L_0\|\widehat W_k-\widehat W_{k-\tau}\|_2]\leq2\gamma L_0\eta\tau G_{\max}.
\end{aligned}
\]
}
For the drift terms, Young's inequality gives

{\color{black}
Here $\mathbb E[Z_{k+1}\mid\mathcal I_k]=\gamma$.
}
{\color{black}
Both $e_k$ and $\widehat W_k-\widehat W_{k-\tau}$ are $\mathcal I_k$-measurable.
Young's inequality therefore gives
\[
\begin{aligned}
&-2\eta\mathbb E\Big\langle
\Bigl(g_{0,k}(\widehat W_k)-g_{0,k}(\widehat W_{k-\tau})\Bigr)
+\gamma\Bigl(\bar g_0^{\mathrm{POMDP}}(\widehat W_{k-\tau})-\bar g_0^{\mathrm{POMDP}}(\widehat W_k)\Bigr),e_k\Big\rangle\\
&\leq4\eta\gamma L_0\mathbb E\!\left[\|\widehat W_k-\widehat W_{k-\tau}\|_2\|e_k\|_2\right]\\
&\leq\eta\gamma\alpha\mathbb E\|e_k\|_2^2
+\frac{4\eta\gamma L_0^2}{\alpha}\mathbb E\|\widehat W_k-\widehat W_{k-\tau}\|_2^2
\leq\eta\gamma\alpha\mathbb E\|e_k\|_2^2
+\frac{4\gamma L_0^2}{\alpha}\eta^3\tau^2G_{\max}^2.
\end{aligned}
\]
}

\subparagraph{Step 5.2. Bound on lagged bias.}

For the lagged bias term in \eqref{eq:zeta-lagged-decomp-pre-reset}, we freeze the iterate at   {\color{black}
$\widehat W_{k-\tau}$ and condition on
} $\mathcal I_{k-\tau}$.

{\color{black}
Then
}

{\color{black}
\[
\begin{aligned}
&g_{0,k}(\widehat W_{k-\tau})-\gamma\bar g_0^{\mathrm{POMDP}}(\widehat W_{k-\tau})\\
&=\underbrace{\Bigl(g_{0,k}(\widehat W_{k-\tau})-\mathbb E[g_{0,k}(\widehat W_{k-\tau})\mid\mathcal I_{k-\tau}]\Bigr)}_{\Xi_{k,\tau}\text{ (martingale difference)}}\\
&\quad+\underbrace{\Bigl(\mathbb E[g_{0,k}(\widehat W_{k-\tau})\mid\mathcal I_{k-\tau}]-\gamma\bar g_0^{\mathrm{POMDP}}(\widehat W_{k-\tau})\Bigr)}_{\Gamma_{k,\tau}^{\mathrm{mix}}\text{ (mixing bias)}}.
\end{aligned}
\]
}

{\color{black}
The population mismatch is already included in $\omega_k$.
}

Since $\mathbb E[\Xi_{k,\tau}\mid\mathcal I_{k-\tau}]=0$, the term simplifies to
\[
\mathbb E\langle\Xi_{k,\tau},e_k\rangle=\mathbb E\langle\Xi_{k,\tau},e_k-e_{k-\tau}\rangle.
\]

{\color{black}
We have $\|\Xi_{k,\tau}\|_2\leq2G_{\max}$ and $\|e_k-e_{k-\tau}\|_2\leq\eta\tau G_{\max}$.
}
{\color{black}
The reset coin also gives $\mathbb E\|\Xi_{k,\tau}\|_2\leq2\gamma G_{\max}$.
}
{\color{black}
Therefore,
}

{\color{black}
\[
\begin{aligned}
-2\eta\mathbb E\langle\Xi_{k,\tau},e_k\rangle
&=-2\eta\mathbb E\langle\Xi_{k,\tau},e_k-e_{k-\tau}\rangle\\
&\leq2\eta\mathbb E[\|\Xi_{k,\tau}\|_2\|e_k-e_{k-\tau}\|_2]
\leq4\eta^2\gamma\tau G_{\max}^2.
\end{aligned}
\]
}
{\color{black}
The expectation bound follows from $\|g_{0,k}(W)\|_2\leq Z_{k+1}G_{\max}$ and conditional Jensen's inequality.
}

\textbf{Bound on $\Gamma_{k,\tau}^{\mathrm{mix}}$.}

{\color{black}
Given $X_k$, the mean of $g_{0,k}(W)$ is bounded by $\gamma G_{\max}$ for every fixed $W$.
}

{\color{black}
For the actual augmented POMDP chain, Lemma~\ref{lem:actual-reset} gives
}

{\color{black}
\[
\|\mathbb P(X_k\in\cdot\mid\mathcal I_{k-\tau})-\xi^\pi\|_{\mathrm{TV}}
\leq \gamma^\tau.
\]
}

{\color{black}
Hence,
}

{\color{black}
\[
\|\Gamma_{k,\tau}^{\mathrm{mix}}\|_2
\leq2\gamma G_{\max}\|\mathbb P(X_k\in\cdot\mid\mathcal I_{k-\tau})-\xi^\pi\|_{\mathrm{TV}}
\leq2\gamma G_{\max}\gamma^\tau.
\]
}
Applying Young's inequality for any $\alpha > 0$, the inner product is bounded by

{\color{black}
\[
-2\eta\mathbb E\langle\Gamma_{k,\tau}^{\mathrm{mix}},e_k\rangle
\leq2\eta\mathbb E[\|\Gamma_{k,\tau}^{\mathrm{mix}}\|_2\|e_k\|_2]
\leq\eta\gamma\alpha\mathbb E\|e_k\|_2^2
+\frac{4\eta\gamma G_{\max}^2\gamma^{2\tau}}{\alpha}.
\]
}

{\color{black}
Combining the drift, martingale, and mixing bounds gives
}

{\color{black}
\[
-2\eta\mathbb E\langle\zeta_k,e_k\rangle
\leq2\eta\gamma\alpha\mathbb E\|e_k\|_2^2
+\frac{4\gamma L_0^2}{\alpha}\eta^3\tau^2G_{\max}^2
+4\eta^2\gamma\tau G_{\max}^2
+\frac{4\eta\gamma G_{\max}^2\gamma^{2\tau}}{\alpha}.
\]
}

\paragraph{\textbf{Step 6. Combine all}.}

{\color{black}
Taking expectation in \eqref{eq:lyapunov-recursion-decomp-pre-reset} and combining the bounds from Steps 2--5, we obtain the following inequality for every $k\geq\tau$.
}

{\color{black}
\[
\begin{aligned}
&2\eta\gamma(1-\sqrt\gamma)\mathbb E\|\widehat Q_0(\cdot,\cdot;\widehat W_k)-Q_{\mathcal F}^{*,\pi}\|_{\mu^\pi}^2\leq(1+5\eta\gamma\alpha)\mathbb E\|e_k\|_2^2-\mathbb E\|e_{k+1}\|_2^2\\
&\quad+\frac{\eta\gamma}{\alpha}O\!\left(\delta_{\max}^2R^{2/3}m^{-1/3}d^5\log m+R^{8/3}m^{-1/3}d^8\log m
+\frac{L_Q^2\delta_{\max}^2}{(1-\gamma)^2}(1-\rho)^{2\ell}\right)\\
&\quad+\frac{4\gamma L_0^2}{\alpha}\eta^3\tau^2G_{\max}^2
+4\eta^2\gamma\tau G_{\max}^2
+\frac{4\eta\gamma G_{\max}^2\gamma^{2\tau}}{\alpha}
+\eta^2\gamma G_{\max}^2.
\end{aligned}
\]
}

{\color{black}
We sum the preceding inequality from $k=\tau$ to $K-1$. The Lyapunov terms telescope as
\[
\sum_{k=\tau}^{K-1}\left(\mathbb E\|e_k\|_2^2-\mathbb E\|e_{k+1}\|_2^2\right)
=\mathbb E\|e_\tau\|_2^2-\mathbb E\|e_K\|_2^2
\leq4dR^2.
\]
Moreover, $\|e_k\|_2^2\leq4dR^2$ implies
\[
5\eta\gamma\alpha\sum_{k=\tau}^{K-1}\mathbb E\|e_k\|_2^2
\leq20\eta\gamma\alpha dR^2(K-\tau).
\]
Algorithm~\ref{alg:neural_superstate_td} averages
$\widehat W_0,\ldots,\widehat W_K$. Since $\widehat Q_0$ is affine in $W$, Jensen's inequality gives
\[
\mathbb E\|\widehat Q_0(\bar W_K)-Q_{\mathcal F}^{*,\pi}\|_{\mu^\pi}^2
\leq\frac{1}{K+1}\sum_{k=0}^{K}
\mathbb E\|\widehat Q_0(\widehat W_k)-Q_{\mathcal F}^{*,\pi}\|_{\mu^\pi}^2.
\]
The first $\tau$ iterates and the last iterate have squared projected function error at most $4Q_{\max}^2$. Therefore, for an absolute constant $C$,
\[
\begin{aligned}
&\mathbb E\|\widehat Q_0(\bar W_K)-Q_{\mathcal F}^{*,\pi}\|_{\mu^\pi}^2\\
&\leq\frac{C}{1-\sqrt\gamma}\left(\frac{dR^2}{\gamma\eta K}+dR^2\alpha+\eta(\tau+1)G_{\max}^2\right)\\
&\quad+\frac{1}{\alpha(1-\sqrt\gamma)}O\!\left(
\delta_{\max}^2R^{2/3}m^{-1/3}d^5\log m+R^{8/3}m^{-1/3}d^8\log m
+\frac{L_Q^2\delta_{\max}^2}{(1-\gamma)^2}(1-\rho)^{2\ell}\right)\\
&\quad+\frac{C L_0^2\eta^2\tau^2G_{\max}^2+C G_{\max}^2\gamma^{2\tau}}
{\alpha(1-\sqrt\gamma)}
+\frac{4(\tau+1)Q_{\max}^2}{K+1}.
\end{aligned}
\]

Choose
\[
\eta=K^{-1/2},
\qquad
\tau=\max\left\{1,\left\lceil\frac{\log K}{2|\log\gamma|}\right\rceil\right\},
\qquad K>\tau.
\]
Then $\gamma^\tau\leq K^{-1/2}$. We use the same Young parameter throughout the preceding bounds and set
\[
\begin{aligned}
\alpha=\frac{1}{\sqrt d\,R}\Bigl(&
\delta_{\max}^2R^{2/3}m^{-1/3}d^5\log m+R^{8/3}m^{-1/3}d^8\log m\\
&+\frac{L_Q^2\delta_{\max}^2}{(1-\gamma)^2}(1-\rho)^{2\ell}
+L_0^2\eta^2\tau^2G_{\max}^2+G_{\max}^2\gamma^{2\tau}\Bigr)^{1/2}.
\end{aligned}
\]
This choice balances the $dR^2\alpha$ term with the terms divided by $\alpha$. Using $G_{\max}=L_Q\delta_{\max}$ and the subadditivity of the square root gives
\[
\begin{aligned}
&\mathbb E\|\widehat Q_0(\bar W_K)-Q_{\mathcal F}^{*,\pi}\|_{\mu^\pi}^2\\
&\leq\frac{C}{1-\sqrt\gamma}
\left(\frac{dR^2}{\gamma\eta K}+\eta(\tau+1)G_{\max}^2\right)\\
&\quad+\frac{O\!\left(\delta_{\max}R^{4/3}m^{-1/6}d^3\log^{1/2}m
+R^{7/3}m^{-1/6}d^{9/2}\log^{1/2}m\right)}{1-\sqrt\gamma}\\
&\quad+\frac{C\sqrt d\,R G_{\max}}{1-\sqrt\gamma}
\left(\frac{(1-\rho)^\ell}{1-\gamma}+L_0\eta\tau+\gamma^\tau\right)
+\frac{4(\tau+1)Q_{\max}^2}{K+1}.
\end{aligned}
\]
Finally, $\delta_{\max}=r_{\max}+(1+\gamma)Q_{\max}$ and
$Q_{\max}=O(\log m+d^{3/2}R)$. Thus,
$\delta_{\max}=O(\log m+d^{3/2}R)$ for fixed $r_{\max}$. Substituting these relations together with $L_Q=O(d)$ and $L_0=O(d^2)$ yields
}
{\color{black}
\[
\begin{aligned}
&\mathbb E\!\left[\bigl\|\widehat Q_0(\cdot,\cdot;\bar W_K)-Q_{\mathcal F}^{*,\pi}\bigr\|_{\mu^\pi}^2\right]\\
&\le
\widetilde O\!\left(\frac{d^2+d^{7/2}R+d^5R^2}{\sqrt K}\right)
+\widetilde O\!\left(R^{4/3}m^{-1/6}d^3+R^{7/3}m^{-1/6}d^{9/2}\right)\\
&\quad
+\widetilde O\!\left((d^{3/2}R+d^3R^2)(1-\rho)^\ell\right)
+\widetilde O\!\left(\frac{d^{3/2}R+d^3R^2}{\sqrt K}\right).
\end{aligned}
\]
}

\subsubsection{Bound Approximation Error}

There exists $W^\dagger\in\mathcal B$ such that
\[
\|\widehat Q(\cdot,\cdot;W^\dagger)-\tilde Q^\pi\|_{\mu^\pi}\le \widetilde O\!\left(m^{-2/p}d^{-2/p}\right).
\]
We first decompose
\[
\begin{aligned}
\|Q_{\mathcal F}^{*,\pi}-\tilde Q^\pi\|_{\mu^\pi}
&\le \|Q_{\mathcal F}^{*,\pi}-\widehat Q_0(\cdot,\cdot;W^\dagger)\|_{\mu^\pi}
+\|\widehat Q_0(\cdot,\cdot;W^\dagger)-\widehat Q(\cdot,\cdot;W^\dagger)\|_{\mu^\pi}
\\&\qquad+\|\widehat Q(\cdot,\cdot;W^\dagger)-\tilde Q^\pi\|_{\mu^\pi}.
\end{aligned}
\]
For the first term, we have
\[
\begin{aligned}
&\|Q_{\mathcal F}^{*,\pi}-\widehat Q_0(\cdot,\cdot;W^\dagger)\|_{\mu^\pi}
\le \|Q_{\mathcal F}^{*,\pi}-\Pi_{\mathcal F}\tilde Q^\pi\|_{\mu^\pi}
+\|\Pi_{\mathcal F}\tilde Q^\pi-\widehat Q_0(\cdot,\cdot;W^\dagger)\|_{\mu^\pi}\\
&\qquad\le \|\Pi_{\mathcal F}\tilde T^\pi Q_{\mathcal F}^{*,\pi}-\Pi_{\mathcal F}\tilde T^\pi\tilde Q^\pi\|_{\mu^\pi}
+\|\widehat Q_0(\cdot,\cdot;W^\dagger)-\tilde Q^\pi\|_{\mu^\pi}\\
&\qquad\le
\sqrt\gamma\|Q_{\mathcal F}^{*,\pi}-\tilde Q^\pi\|_{\mu^\pi}
+\|\widehat Q_0(\cdot,\cdot;W^\dagger)-\widehat Q(\cdot,\cdot;W^\dagger)\|_{\mu^\pi}
+\|\widehat Q(\cdot,\cdot;W^\dagger)-\tilde Q^\pi\|_{\mu^\pi}.
\end{aligned}
\]
Here, the first inequality is by the triangle inequality; the second uses $Q_{\mathcal F}^{*,\pi}=\Pi_{\mathcal F}\tilde T^\pi Q_{\mathcal F}^{*,\pi}$, $\tilde Q^\pi=\tilde T^\pi\tilde Q^\pi$, and the projection property $\|\Pi_{\mathcal F}\tilde Q^\pi-\widehat Q_0(\cdot,\cdot;W^\dagger)\|_{\mu^\pi}\le \|\tilde Q^\pi-\widehat Q_0(\cdot,\cdot;W^\dagger)\|_{\mu^\pi}$ since $\widehat Q_0(\cdot,\cdot;W^\dagger)\in\mathcal F$; the third follows from the non-expansiveness of $\Pi_{\mathcal F}$, the
$\sqrt\gamma$-contraction of $\tilde T^\pi$, and the triangle inequality.

Substituting this bound into the initial decomposition gives
\[
\begin{aligned}
\|Q_{\mathcal F}^{*,\pi}-\tilde Q^\pi\|_{\mu^\pi}
&\le
\sqrt\gamma\|Q_{\mathcal F}^{*,\pi}-\tilde Q^\pi\|_{\mu^\pi}
+2\|\widehat Q_0(\cdot,\cdot;W^\dagger)-\widehat Q(\cdot,\cdot;W^\dagger)\|_{\mu^\pi}
\\&\qquad+2\|\widehat Q(\cdot,\cdot;W^\dagger)-\tilde Q^\pi\|_{\mu^\pi}.
\end{aligned}
\]
Rearranging yields
\[
\|Q_{\mathcal F}^{*,\pi}-\tilde Q^\pi\|_{\mu^\pi}
\le
\frac{2}{1-\sqrt\gamma}\left(
\|\widehat Q_0(\cdot,\cdot;W^\dagger)-\widehat Q(\cdot,\cdot;W^\dagger)\|_{\mu^\pi}
+\|\widehat Q(\cdot,\cdot;W^\dagger)-\tilde Q^\pi\|_{\mu^\pi}\right).
\]
Finally, by Fact~\ref{fact:local-linearization} 
\[
\|\widehat Q_0(\cdot,\cdot;W^\dagger)-\widehat Q(\cdot,\cdot;W^\dagger)\|_{\mu^\pi}
\le O\!\left(R^{4/3}m^{-1/6}d^3\log^{1/2}m\right),
\]
and
\[
\|\widehat Q(\cdot,\cdot;W^\dagger)-\tilde Q^\pi\|_{\mu^\pi}
\le \widetilde O\!\left(m^{-2/p}d^{-2/p}\right).
\]
Therefore,
\[
\|Q_{\mathcal F}^{*,\pi}-\tilde Q^\pi\|_{\mu^\pi}
\le \widetilde O\!\left(
\frac{R^{4/3}m^{-1/6}d^3+m^{-2/p}d^{-2/p}}{1-\sqrt\gamma}\right).
\]
Hence,
\[
\|Q_{\mathcal F}^{*,\pi}-\tilde Q^\pi\|_{\mu^\pi}^2
\le \widetilde O\!\left(
\frac{\big(R^{4/3}m^{-1/6}d^3+m^{-2/p}d^{-2/p}\big)^2}{(1-\sqrt\gamma)^2}\right).
\]

Combining this bound with the NTK linearization bound and the projected optimization bound, yields Theorem~\ref{thm:critic-error}.

\section{Proof of Policy Update}
\label{policy improvement}

\subsection{Preliminaries}

First, we give the equivalence of policy update forms for analysis, where we use Recursive Softmax Form in Algorithm \ref{alg:superstate_policy_opt}.
\begin{lemma}[Equivalence of Policy Update Forms]
\label{lem:update-equivalence}
For a fixed superstate 
$B \in \mathsf B$
 and a stepsize $\kappa > 0$, starting from a uniform initial policy $\pi_0(\cdot|B)$, the following three update rules are equivalent for all $t \ge 0$:
\begin{enumerate}
\item
\textbf{Mirror Descent Form:}
\begin{equation}
        \pi_{t+1}(\cdot|B) = \arg\min_{p \in \Delta(\mathcal{A})} \left\{ -\kappa \langle p, \widehat{Q}_t(B,\cdot) \rangle + D_{\mathrm{KL}}(p \| \pi_t(\cdot|B)) \right\}. \label{eq:mirror-descent}
    \end{equation}
\item
\textbf{Recursive Softmax Form:}
\begin{equation}
        \pi_{t+1}(a|B) = \frac{\pi_t(a|B) \exp\bigl(\kappa \widehat{Q}_t(B,a)\bigr)}{\sum_{a' \in \mathcal{A}} \pi_t(a'|B) \exp\bigl(\kappa \widehat{Q}_t(B,a')\bigr)}. \label{eq:recursive-softmax}
    \end{equation}
\item
\textbf{Cumulative Exponential-Weights Form:}
\begin{equation}
        \pi_{t+1}(a|B) = \frac{\exp\bigl(\kappa \sum_{j=0}^t \widehat{Q}_j(B,a)\bigr)}{\sum_{a' \in \mathcal{A}} \exp\bigl(\kappa \sum_{j=0}^t \widehat{Q}_j(B,a')\bigr)}. \label{eq:cumulative-hedge}
    \end{equation}
\end{enumerate}
\end{lemma}

\begin{proof}
Fix any superstate  
$B \in \mathsf B$
.
First, to show equivalence between \eqref{eq:mirror-descent} and \eqref{eq:recursive-softmax}, we consider the Lagrangian of the Mirror Descent objective with multiplier $\lambda$ for the simplex constraint $\sum_a p(a) = 1$:
\[
\mathcal{L}(p, \lambda) = -\kappa \sum_a p(a)\widehat{Q}_t(B,a) + \sum_a p(a)\log\frac{p(a)}{\pi_t(a|B)} + \lambda\Bigl(\sum_a p(a)-1\Bigr).
\]
Setting the first-order optimality condition $\frac{\partial \mathcal{L}}{\partial p(a)} = -\kappa \widehat{Q}_t(B,a) + \log\frac{p(a)}{\pi_t(a|B)} + 1 + \lambda = 0$ yields the characterization:
\begin{equation}
    \pi_{t+1}(a|B) = \pi_t(a|B) \exp(\kappa \widehat{Q}_t(B,a) - 1 - \lambda) \propto \pi_t(a|B) \exp(\kappa \widehat{Q}_t(B,a)).
\end{equation}
Normalizing over $\mathcal{A}$ recovers the recursive softmax update \eqref{eq:recursive-softmax}.

Next, the equivalence between the recursive form \eqref{eq:recursive-softmax} and cumulative form \eqref{eq:cumulative-hedge} follows by induction.
For the base case $t=0$, the uniform initialization $\pi_0(a|B) = 1/|\mathcal{A}|$ directly gives $\pi_1(a|B) \propto \exp(\kappa \widehat{Q}_0(B,a))$.
Assuming the equivalence holds for $\pi_t$, then for $\pi_{t+1}$ we have the chained relation:
\begin{equation}
    \pi_{t+1}(a|B) \propto \pi_t(a|B) e^{\kappa \widehat{Q}_t(B,a)} \propto e^{\kappa \sum_{j=0}^{t-1} \widehat{Q}_j(B,a)} \cdot e^{\kappa \widehat{Q}_t(B,a)} = e^{\kappa \sum_{j=0}^t \widehat{Q}_j(B,a)},
\end{equation}
where the second proportionality follows from the induction hypothesis.
Normalization recovers the cumulative exponential-weights form, completing the proof.
\end{proof}

\paragraph{The Goal.}

Let $N:=TK$%
{\color{black}.}
{\color{black}
Let $\pi^*$ denote an unrestricted optimal policy of the original POMDP.
}
{\color{black}
Let $\widetilde\pi^*$ denote an optimal policy of the fixed superstate MDP.
}
{\color{black}
Following \citet{anjarlekar2025scalable}, we compare the unrestricted POMDP optimum with the surrogate value achieved by each learned superstate policy.
}
{\color{black}
Starting from $H_0=\varnothing$ and $B_0=\mathcal G(H_0)=\varnothing$, define
}
{\color{black}
\[
R_N
:=
K\sum_{t=0}^{T-1}
\mathbb E\!\left[
V^{\pi^*}(H_0)-\tilde V^{\pi_t}(B_0)
\right].
\]
}
We can decompose
{\color{black}
\[
\begin{aligned}
R_N
&=
\underbrace{K\sum_{t=0}^{T-1}\mathbb E\bigl[V^{\pi^*}(H_0)-\tilde V^{\widetilde\pi^*}(B_0)\bigr]}_{\text{model approximation error}}
+
\underbrace{K\sum_{t=0}^{T-1}\mathbb E\bigl[\tilde V^{\widetilde\pi^*}(B_0)-\tilde V^{\pi_t}(B_0)\bigr]}_{\text{policy update regret}}.
\end{aligned}
\]
}
\paragraph*{\color{black}Bound Model Approximation Error.}

{\color{black}
By Theorem~\ref{thm:finite-memory-near-optimality}, we have
}
{\color{black}
\[
\left|V^{\pi^*}(H_0)-\tilde V^{\widetilde\pi^*}(B_0)\right|
\leq
\frac{2r_{\max}}{(1-\gamma)^2}(1-\rho)^\ell.
\]
}
Therefore, the model approximation part of the regret satisfies
{\color{black}
\[
K\sum_{t=0}^{T-1}\mathbb E\bigl[V^{\pi^*}(H_0)-\tilde V^{\widetilde\pi^*}(B_0)\bigr]
\leq
\frac{2Nr_{\max}}{(1-\gamma)^2}(1-\rho)^\ell.
\]
}

\subsection{Bound Superstate Policy Regret}

We now bound the superstate policy-regret term
{\color{black}
$
K\sum_{t=0}^{T-1}\mathbb E\bigl[\tilde V^{\widetilde\pi^*}(B_{0})-\tilde V^{\pi_t}(B_{0})\bigr].
$
}

 {\color{black}
\ Starting from $B_0=\varnothing$
}, define the normalized discounted visitation distribution under policy $\pi$ by
\[
d^\pi(B):=(1-\gamma)\sum_{j=0}^\infty \gamma^j \mathbb P(B_j=B\mid B_0,\pi).
\]
{\color{black}
Under the full-reset SMDP construction, $d^\pi$ coincides with the stationary superstate distribution by Lemma~\ref{lem:occupancy-stationarity}. Hence $\mu^\pi$ has marginal $d^\pi$.
}
{\color{black}
These are the discounted occupancies of the original SMDP.
}

\paragraph{\textbf{Step 1. Performance-difference decomposition.}}

By the performance difference lemma, for every $t$,
{\color{black}
\[
\tilde V^{\widetilde{\pi}^*}(B_{0})-\tilde V^{\pi_t}(B_{0})
=
\frac{1}{1-\gamma}
\sum_{B'\in\mathsf B}d^{\widetilde{\pi}^*}(B')
\sum_{a\in\mathcal A}
\bigl(\widetilde{\pi}^*(a\mid B')-\pi_t(a\mid B')\bigr)
\tilde Q^{\pi_t}(B',a).
\]
}
Summing over $t$ and multiplying by $K$, and then adding and subtracting
$\widehat Q_t$, we obtain
{\color{black}
\[
\begin{aligned}
&K\sum_{t=0}^{T-1}\mathbb E\bigl[\tilde V^{\widetilde{\pi}^*}(B_{0})-\tilde V^{\pi_t}(B_{0})\bigr]\\
&=
\frac{K}{1-\gamma}\sum_{t=0}^{T-1}
\mathbb E\Biggl[
\sum_{B,a}d^{\widetilde{\pi}^*}(B)
\bigl(\widetilde{\pi}^*(a\mid B)-\pi_t(a\mid B)\bigr)
\widehat Q_t(B,a)
\Biggr]\\
&\quad-
\frac{K}{1-\gamma}\sum_{t=0}^{T-1}
\mathbb E\Biggl[
\sum_{B,a}d^{\widetilde{\pi}^*}(B)
\bigl(\widetilde{\pi}^*(a\mid B)-\pi_t(a\mid B)\bigr)
\bigl(\widehat Q_t(B,a)-\tilde Q^{\pi_t}(B,a)\bigr)
\Biggr].
\end{aligned}
\]
}
The first term is the PMD optimization error with respect to the estimated critic, and the second term is the critic-induced error.

\paragraph{\textbf{Step 2. PMD optimization error.}}

For each fixed superstate $B$, the recursive softmax update is the entropic mirror-descent update on the simplex:
\[
\pi_{t+1}(\cdot\mid B)
=
\arg\min_{p\in\Delta(\mathcal A)}
\left\{
-\kappa\langle p,\widehat Q_t(B,\cdot)\rangle
+D_{\mathrm{KL}}(p\|\pi_t(\cdot\mid B))
\right\}.
\]
Therefore, the standard exponential-weights regret bound gives, for comparator
 {\color{black}
$\widetilde{\pi}^*(\cdot\mid B)$
},
{\color{black}
\[
\sum_{t=0}^{T-1}
\langle
\widetilde{\pi}^*(\cdot\mid B)-\pi_t(\cdot\mid B),
\widehat Q_t(B,\cdot)
\rangle
\le
\frac{D_{\mathrm{KL}}(\widetilde{\pi}^*(\cdot\mid B)\|\pi_0(\cdot\mid B))}{\kappa}
+
\frac{\kappa}{2}\sum_{t=0}^{T-1}
\|\widehat Q_t(B,\cdot)\|_\infty^2.
\]
}
Since $\pi_0(\cdot\mid B)$ is uniform and
$\|\widehat Q_t(B,\cdot)\|_\infty\le Q_{\max}$, we have
{\color{black}
\[
\sum_{t=0}^{T-1}
\langle
\widetilde{\pi}^*(\cdot\mid B)-\pi_t(\cdot\mid B),
\widehat Q_t(B,\cdot)
\rangle
\le
\frac{\log|\mathcal A|}{\kappa}
+
\frac{\kappa TQ_{\max}^2}{2}.
\]
}
Taking expectation over  %
 {\color{black}
$B\sim d^{\widetilde{\pi}^*}$
} gives
{\color{black}
\[
\sum_{t=0}^{T-1}
\mathbb E_{B\sim d^{\widetilde{\pi}^*}}
\bigl[
\langle
\widetilde{\pi}^*(\cdot\mid B)-\pi_t(\cdot\mid B),
\widehat Q_t(B,\cdot)
\rangle
\bigr]
\le
\frac{\log|\mathcal A|}{\kappa}
+
\frac{\kappa TQ_{\max}^2}{2}.
\]
}

{\color{black}
Taking expectation over algorithmic randomness preserves this upper bound.
}

\paragraph{\textbf{Step 3. Critic-induced error under concentrability.}}

Conditioning on $\pi_t$, by the triangle inequality
 {\color{black}
$|\widetilde{\pi}^*(a\mid B)-\pi_t(a\mid B)|\le \widetilde{\pi}^*(a\mid B)+\pi_t(a\mid B)$
}, we have
{\color{black}
\[
\begin{aligned}
&\left|
\sum_{B,a}d^{\widetilde{\pi}^*}(B)
\bigl(\widetilde{\pi}^*(a\mid B)-\pi_t(a\mid B)\bigr)
\bigl(\widehat Q_t(B,a)-\tilde Q^{\pi_t}(B,a)\bigr)
\right|\\
&\le
\sum_{B,a}
\bigl(\mu^{\widetilde{\pi}^*}(B,a)+d^{\widetilde{\pi}^*}(B)\pi_t(a\mid B)\bigr)
\bigl|
\widehat Q_t(B,a)-\tilde Q^{\pi_t}(B,a)
\bigr|.
\end{aligned}
\]
}
For the first term, Cauchy--Schwarz gives
{\color{black}
\[
\sum_{B,a}\mu^{\widetilde{\pi}^*}(B,a)
\bigl|\widehat Q_t-\tilde Q^{\pi_t}\bigr|
\le
C_t^{1/2}
\|\widehat Q_t-\tilde Q^{\pi_t}\|_{\mu^{\pi_t}}.
\]
}
For the second term, since
 {\color{black}
$d^{\widetilde{\pi}^*}(B)\pi_t(a\mid B)
=
\mu^{\pi_t}(B,a)d^{\widetilde{\pi}^*}(B)/d^{\pi_t}(B)$
}, another application of Cauchy--Schwarz yields
{\color{black}
\[
\sum_{B,a}d^{\widetilde{\pi}^*}(B)\pi_t(a\mid B)
\bigl|\widehat Q_t-\tilde Q^{\pi_t}\bigr|
\le
\left(\sum_B\frac{(d^{\widetilde{\pi}^*}(B))^2}{d^{\pi_t}(B)}\right)^{1/2}
\|\widehat Q_t-\tilde Q^{\pi_t}\|_{\mu^{\pi_t}}.
\]
}
Moreover,
{\color{black}
\[
\sum_B\frac{(d^{\widetilde{\pi}^*}(B))^2}{d^{\pi_t}(B)}
\le
\sum_{B,a}\frac{(\mu^{\widetilde{\pi}^*}(B,a))^2}{\mu^{\pi_t}(B,a)}
=
C_t.
\]
}
Thus,
{\color{black}
\[
\left|
\sum_{B,a}d^{\widetilde{\pi}^*}(B)
\bigl(\widetilde{\pi}^*(a\mid B)-\pi_t(a\mid B)\bigr)
\bigl(\widehat Q_t(B,a)-\tilde Q^{\pi_t}(B,a)\bigr)
\right|
\le
2C_t^{1/2}
\|\widehat Q_t-\tilde Q^{\pi_t}\|_{\mu^{\pi_t}}.
\]
}
Taking expectation and applying Cauchy--Schwarz, Assumption~\ref{assum:coverage}, and Theorem~\ref{thm:critic-error}, we obtain
{\color{black}
\[
\begin{aligned}
&\mathbb E\left[
\left|
\sum_{B,a}d^{\widetilde\pi^*}(B)
\bigl(\widetilde\pi^*(a\mid B)-\pi_t(a\mid B)\bigr)
\bigl(\widehat Q_t(B,a)-\tilde Q^{\pi_t}(B,a)\bigr)
\right|\right]\\
&\leq2\mathbb E\left[C_t^{1/2}\|\widehat Q_t-\tilde Q^{\pi_t}\|_{\mu^{\pi_t}}\right]\leq2\sqrt{\mathbb E[C_t]}\,
\sqrt{\mathbb E\left[\|\widehat Q_t-\tilde Q^{\pi_t}\|_{\mu^{\pi_t}}^2\right]}
\leq2\sqrt{\bar C_\infty\varepsilon_K}.
\end{aligned}
\]
}
where $\varepsilon_K=\mathbb E\|\widehat Q(\cdot,\cdot;\bar W_K)-\tilde Q^\pi\|_{\mu^\pi}^2$ denotes the squared critic-error upper bound from Theorem~\ref{thm:critic-error}.

\paragraph{\textbf{Step 4. Combine the two bounds.}}

Combining Steps 1--3 gives
{\color{black}
\[
K\sum_{t=0}^{T-1}\mathbb E\bigl[\tilde V^{\widetilde{\pi}^*}(B_{0})-\tilde V^{\pi_t}(B_{0})\bigr]
\le
\frac{K}{1-\gamma}
\left(
\frac{\log|\mathcal A|}{\kappa}
+
\frac{\kappa TQ_{\max}^2}{2}
+
2T\sqrt{\bar C_\infty\varepsilon_K}
\right).
\]
}

Together with the model-approximation bound,
{\color{black}
\[
K\sum_{t=0}^{T-1}
\mathbb E\bigl[V^{\pi^*}(H_{0})-\tilde V^{\widetilde{\pi}^*}(B_{0})\bigr]
\le
\frac{2Nr_{\max}}{(1-\gamma)^2}(1-\rho)^\ell,
\]
}
we obtain
\[
R_N
\le
\frac{2Nr_{\max}}{(1-\gamma)^2}(1-\rho)^\ell
+
\frac{K}{1-\gamma}
\left(
\frac{\log|\mathcal A|}{\kappa}
+
\frac{\kappa TQ_{\max}^2}{2}
+
2T\sqrt{\bar C_\infty\varepsilon_K}
\right).
\]
Choosing
$
\kappa=\frac{\sqrt{2\log|\mathcal A|}}{Q_{\max}\sqrt T}
$
balances the two PMD terms and gives
\[
\frac{\log|\mathcal A|}{\kappa}
+
\frac{\kappa TQ_{\max}^2}{2}
=
Q_{\max}\sqrt{2T\log|\mathcal A|}.
\]
Since $N=TK$, this yields
\[
R_N
\le
\frac{2Nr_{\max}}{(1-\gamma)^2}(1-\rho)^\ell
+
\frac{N}{1-\gamma}
\left(
Q_{\max}\sqrt{\frac{2\log|\mathcal A|}{T}}
+
2\sqrt{\bar C_\infty\varepsilon_K}
\right).
\]

\paragraph{{\color{black}\textbf{Step 5. Substitute the critic bound and choose $K,T$.}}}

{\color{black}Each critic run satisfies the conditions of Theorem~\ref{thm:critic-error} as required in Theorem~\ref{thm:regret}.}
By Theorem~\ref{thm:critic-error} and $\sqrt{a_1+\cdots+a_s}\leq\sqrt{a_1}+\cdots+\sqrt{a_s}$,  {\color{black}
we obtain
}

{\color{black}
\[
\begin{aligned}
\sqrt{\varepsilon_K}
&\leq\widetilde O\!\left(m^{-2/p}d^{-2/p}\right)
+\widetilde O\!\left(R^{4/3}m^{-1/6}d^3+R^{2/3}m^{-1/12}d^{3/2}+R^{7/6}m^{-1/12}d^{9/4}\right)\\
&\quad+\widetilde O\!\left(\frac{d+d^{7/4}R^{1/2}+d^{5/2}R}{K^{1/4}}\right)
+\widetilde O\!\left((d^{3/4}R^{1/2}+d^{3/2}R)(1-\rho)^{\ell/2}\right)\\
&\quad+\widetilde O\!\left((d^{3/4}R^{1/2}+d^{3/2}R)\gamma^{\tau/2}\right).
\end{aligned}
\]
}

{\color{black}
Choose
}
$
K=N^{2/3},T=N^{1/3}.
$

{\color{black}
Choose the same analysis lag as above,
$\tau=\max\{1,\lceil\log K/(2|\log\gamma|)\rceil\}$, with $K>\tau$. Then
}
{\color{black}
\[
T^{-1/2}=K^{-1/4}=N^{-1/6},\qquad\gamma^{\tau/2}\leq K^{-1/4}.
\]
}
{\color{black}
The mixing polynomial satisfies $d^{3/4}R^{1/2}+d^{3/2}R\leq P_{\rm opt}(d,R)$ for $d\geq1$.
}

Using also $Q_{\max}=O(\log m+d^{3/2}R)$ and absorbing logarithmic factors into
$\widetilde O(\cdot)$, we obtain  {\color{black}the following bound:}

{\color{black}
\[
\begin{aligned}
&R_N\leq N\Big[
\widetilde O(m^{-2/p}d^{-2/p})
+\widetilde O(R^{4/3}m^{-1/6}d^3+R^{2/3}m^{-1/12}d^{3/2}+R^{7/6}m^{-1/12}d^{9/4})\\
&\qquad\quad+\widetilde O((1-\rho)^\ell)
+\widetilde O((d^{3/4}R^{1/2}+d^{3/2}R)(1-\rho)^{\ell/2})\Big]\\
&\qquad+\widetilde O\!\left(N^{5/6}(1+d+d^{3/2}R+d^{7/4}R^{1/2}+d^{5/2}R)\right).
\end{aligned}
\]
}

\clearpage
\section{Additional Experimental Results}
\label{app:experiments}

\begin{figure}[t]
\centering
\includegraphics[width=0.5\linewidth]{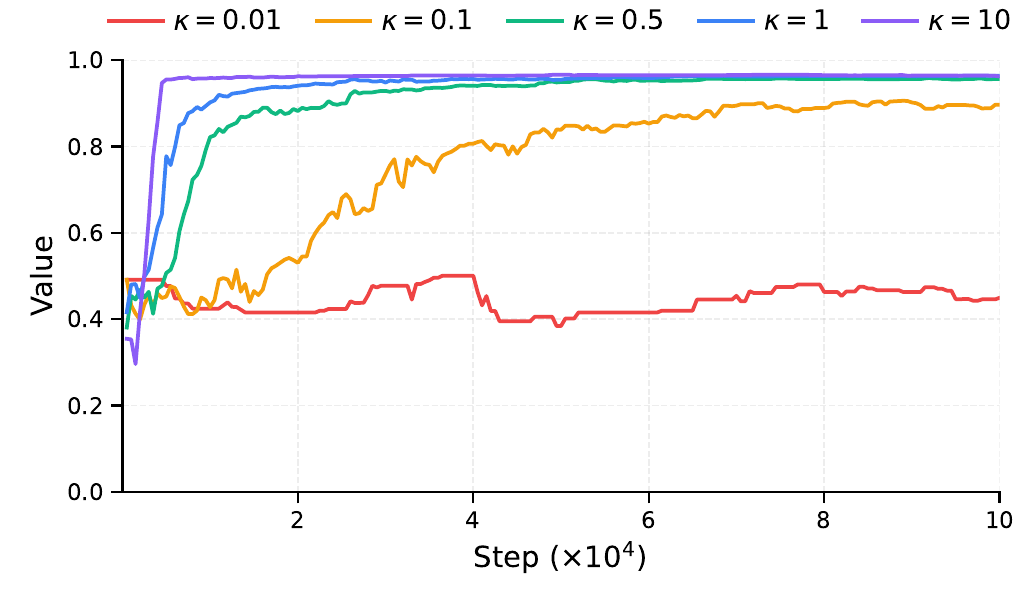}
\caption{\textbf{Sensitivity to the mirror descent parameter $\kappa$ in CheeseMaze.}
The y-axis reports policy value, defined as the Monte Carlo estimate of the mean discounted return $\widehat V=\mathbb{E}[\sum_h \gamma^h r_h]$ over 50 evaluation episodes. Evaluation is performed every 500 training steps. Curves show the mean over 3 random seeds.}
\label{fig:kappa_sensitivity}
\end{figure}

\subsection{Sensitivity to the Policy Stepsize}
\label{app:kappa-sensitivity}

Figure~\ref{fig:kappa_sensitivity} reports the policy value of our neural PMD on CheeseMaze ($\ell = 2$) for $\kappa \in \{0.01, 0.1, 0.5, 1, 10\}$. Performance is robust over $\kappa \in [0.1, 10]$; only the smallest setting $\kappa = 0.01$ leads to a slow update and noticeably degraded performance. We additionally observed (not shown) that significantly larger values of $\kappa$ harm performance, as expected: the mirror-descent step becomes too aggressive and amplifies the inexact critic.

\subsection{Hyperparameters}
\label{app:hyperparameters}

Table~\ref{tab:hyperparameters} summarizes the hyperparameters used in our experiments. All methods share the same superstate input encoding (concatenated one-hot observations and actions) and the same critic architecture; entries marked ``--'' indicate that the corresponding component is not part of the algorithm.

\begin{table}[h]
\centering
\caption{Hyperparameters used in CheeseMaze experiments.}
\label{tab:hyperparameters}
\begin{tabular}{lccc}
\toprule
 & \textbf{Mirror Descent (ours)} & \textbf{Q-learning} & \textbf{Actor-Critic} \\
\midrule
Discount $\gamma$ & $0.99$ & $0.99$ & $0.99$ \\
Outer iterations $T$ & $200$ & $200$ & $200$ \\
Critic steps per iter $K$ & $500$ & $500$ & $500$ \\
Critic learning rate & $10^{-3}$ & $10^{-3}$ & $10^{-3}$ \\
Critic hidden dim & $64$ & $64$ & $64$ \\
Critic hidden layers & $2$ & $2$ & $2$ \\
Actor learning rate & -- & -- & $10^{-3}$ \\
Actor hidden dim & -- & -- & $64$ \\
Actor hidden layers & -- & -- & $1$ \\
Mirror-descent stepsize $\kappa$ & $0.1$ & -- & -- \\
$\epsilon$-greedy schedule & -- & $1.0 \to 0.05$ (linear) & -- \\
Optimizer & Adam & Adam & Adam \\
Random seeds & $\{1,2,3\}$ & $\{1,2,3\}$ & $\{1,2,3\}$ \\
Evaluation episodes & $50$ & $50$ & $50$ \\
Evaluation interval (steps) & $500$ & $500$ & $500$ \\
\bottomrule
\end{tabular}
\end{table}

\subsection{Compute Resources}
\label{app:compute-resources}
All reported experiments were run on a single NVIDIA H200 GPU with 141GB HBM3e memory. The CheeseMaze experiments are lightweight, and each training run typically takes only a few minutes in our implementation. GPU acceleration was used mainly for convenience rather than necessity. The experiments can be run on smaller GPUs and should also be feasible on CPU-only machines with longer runtime. No special storage resources are required, since the CheeseMaze environment is generated procedurally. The full set of reported experiments can be completed within a few hours.

\subsection{Existing Assets and Licenses}
\label{app:licenses}
We acknowledge the CleanRL project~\citep{huang2022cleanrl}, which our implementation builds on for the experimental framework and implementation structure. CleanRL is publicly available and released under the MIT license. Our code is included in the supplemental material. No external datasets are used.

\section{Broader Impacts}
\label{app:broader-impacts}

This work is primarily foundational and aims to improve the understanding of reinforcement learning under partial observability. Potential positive impacts include practical applications such as autonomous driving, robotics, and resource allocation. Potential negative impacts are indirect: if such methods are deployed in autonomous decision-making systems without sufficient validation, they could lead to harmful decisions. In addition, scaling reinforcement learning methods to larger domains may increase computational and energy costs.

\newpage
\clearpage

\typeout{get arXiv to do 4 passes: Label(s) may have changed. Rerun}
\end{document}